\documentclass[10pt,twocolumn,letterpaper]{article}

\usepackage[pagenumbers]{iccv} 

\usepackage[accsupp]{axessibility} 
\usepackage{amssymb}
\usepackage[linesnumbered,ruled,vlined]{algorithm2e}
\usepackage{enumerate}
\usepackage{algorithmic}
\usepackage[dvipsnames]{xcolor}
\usepackage{tikz}  
\usepackage{bm}
\usepackage{pifont}
\usetikzlibrary{fit, calc}
\usepackage{tabularx}
\usepackage{multirow}
\usepackage{booktabs}
\usepackage{wrapfig}
\usepackage{makecell}
\usepackage{float}
\usepackage{stfloats}
\usepackage{blindtext}
\usepackage[accsupp]{axessibility}
\usepackage[utf8]{inputenc}
\usepackage{graphicx}
\usepackage{booktabs}
\usepackage{colortbl}
\usepackage{microtype}
\usepackage[accsupp]{axessibility}
\usepackage{pgfplots}
\pgfplotsset{compat=1.18}
\usepackage[graphicx]{realboxes}
\usepackage{bm}
\usepackage{nicefrac}
\usepackage{multirow}
\usepackage[number-mode=text]{siunitx}
\usepackage{marvosym}
\usepackage{placeins}
\usepackage{xr}
\usepackage{xpatch}
\usepackage{fancyhdr}
\usepackage{setspace}
\usepackage{threeparttable}
\usepackage{pifont}

\usepackage{xspace}

\newcolumntype{+}{>{\global\let\currentrowstyle\relax}}
\newcolumntype{-}{>{\currentrowstyle}}

\newcommand{\tblgray}{gray!75}

\usepackage{trimclip}
\makeatletter
\DeclareRobustCommand{\circbullet}{\mathbin{\vphantom{\circ}\text{\circbullet@}}}
\newcommand{\circbullet@}{%
  \check@mathfonts
  \m@th\ooalign{%
    \clipbox{0 0 0 {\dimexpr\height-\fontdimen22\textfont2}}{$\bullet$}\cr
    $\circ$\cr
  }%
}
\makeatother

\definecolor{mypink}{HTML}{FFEAEA}
\definecolor{myblue}{HTML}{DBF2F9}
\definecolor{tblblue}{HTML}{0801FF}
\definecolor{tblred}{HTML}{FF0000}
\newcommand{\boxit}[2]{
    \tikz[remember picture,overlay] \node (A) {};\ignorespaces
    \tikz[remember picture,overlay]{\node[yshift=4.75pt,fill=#1,opacity=1.0,fit={($(A)+(0,0.15\baselineskip)$)($(A)+(.87\linewidth,-{#2}\baselineskip - 0.25\baselineskip)$)}] {};}\ignorespaces
}

\definecolor{iccvblue}{rgb}{0.21,0.49,0.74}
\usepackage[pagebackref,breaklinks,colorlinks,citecolor=iccvblue]{hyperref}

\def\paperID{12391} 
\def\confName{ICCV}
\def\confYear{2025}

\title{PRO-VPT: Distribution-Adaptive Visual Prompt Tuning via Prompt Relocation}

\author{
Chikai Shang$^{1, 2}$\hspace{20pt}Mengke Li$^{3, 5}$\hspace{20pt}Yiqun Zhang$^{4, 5}$\hspace{20pt}Zhen Chen$^6$\hspace{20pt}Jinlin Wu$^{6, 7}$\\Fangqing Gu$^{4, 5}$\thanks{Corresponding authors.}\hspace{20pt}Yang Lu$^{1, 2, 5}$\footnotemark[1]\hspace{20pt}Yiu-ming Cheung$^5$\\
{\small $^1$Key Laboratory of Multimedia Trusted Perception and Efficient Computing, Ministry of Education of China, Xiamen University}\vspace{-0.2em}\\
{\small $^2$School of Informatics, Xiamen University\hspace{20pt}$^3$Shenzhen University\hspace{20pt}$^4$Guangdong University of Technology}\vspace{-0.2em}\\
{\small $^5$Hong Kong Baptist University\hspace{20pt}$^6$CAIR, HKISI, CAS\hspace{20pt}$^7$MAIS, CASIA}\vspace{-0.2em}\\
{\tt\small ckshang12@gmail.com\hspace{20pt}fqgu@gdut.edu.cn\hspace{20pt}luyang@xmu.edu.cn}\\
}

\begin{document}
\maketitle

\setlength{\textfloatsep}{0.9em}

\begin{abstract}

\vspace{-0.72em}

Visual prompt tuning (VPT), i.e., fine-tuning some lightweight prompt tokens, provides an efficient and effective approach for adapting pre-trained models to various downstream tasks. However, most prior art indiscriminately uses a \textbf{fixed} prompt distribution across different tasks, neglecting the importance of each block varying depending on the task. In this paper, we introduce adaptive distribution optimization (\textbf{ADO}) by tackling two key questions: \ding{172} How to appropriately and formally define ADO, and \ding{173} How to design an adaptive distribution strategy guided by this definition? Through empirical analysis, we first confirm that properly adjusting the distribution significantly improves VPT performance, and further uncover a key insight that a nested relationship exists between ADO and VPT. Based on these findings, we propose a new VPT framework, termed \textbf{PRO-VPT} (iterative Prompt RelOcation-based VPT), which adaptively adjusts the distribution built upon a nested optimization formulation. Specifically, we develop a prompt relocation strategy derived from this formulation, comprising two steps: pruning idle prompts from prompt-saturated blocks, followed by allocating these prompts to the most prompt-needed blocks. By iteratively performing prompt relocation and VPT, our proposal can adaptively learn the optimal prompt distribution in a nested optimization-based manner, thereby unlocking the full potential of VPT. Extensive experiments demonstrate that our proposal significantly outperforms advanced VPT methods, e.g., PRO-VPT surpasses VPT by 1.6 pp and 2.0 pp average accuracy, leading prompt-based methods to state-of-the-art performance on VTAB-1k and FGVC benchmarks. 
The code is available at~\href{https://github.com/ckshang/PRO-VPT}{\texttt{https://github.com/ckshang/PRO-VPT}}.

\end{abstract}

\vspace{-1.36em}
\section{Introduction}

Fine-tuning pre-trained vision models (PVMs)~\cite{khan2022transformers} has proven remarkably effective in adapting to a variety of downstream tasks~\cite{yu2024visual, han2024facing, li2024improving}. However, the computational and storage costs associated with full fine-tuning are prohibitively high, particularly as model sizes continue to grow~\cite{ma2022xprompt, wang2025lora, chen2023understanding}. To overcome these challenges, parameter-efficient fine-tuning (PEFT) methods~\cite{xin2024parameter, xin2025v, hu2022lora, chen2022adaptformer}, such as visual prompt tuning (VPT)~\cite{jia2022visual}, have emerged as more promising alternatives, garnering significant attention.

\begin{figure}[t]
    \centering
    \includegraphics[scale=1.]{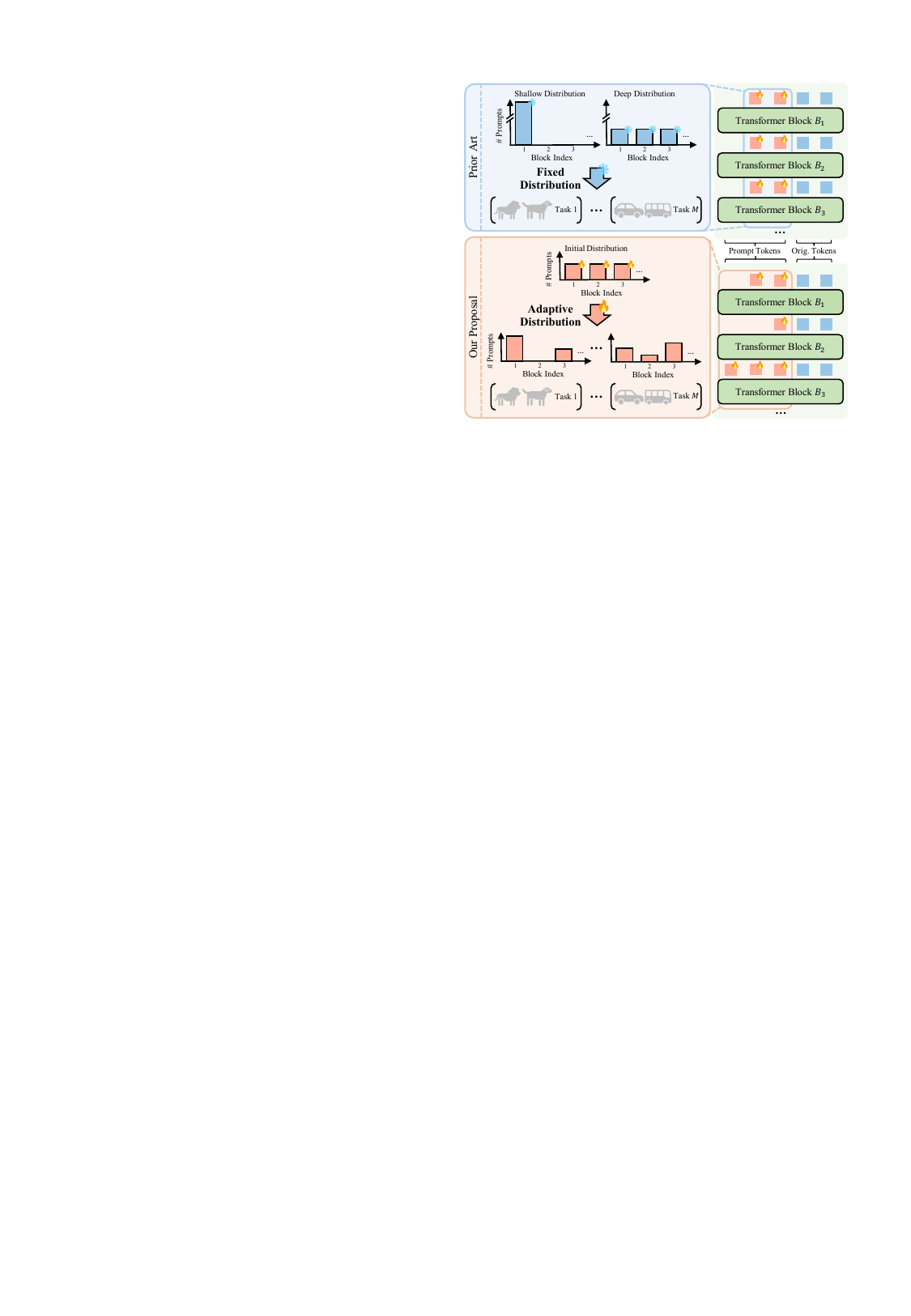}
    \vspace{-0.87em}
    \caption{\textbf{PRO-VPT (ours) vs. prior art in VPT.} Existing VPT approaches typically insert trainable prompts into the PVM with a pre-specific static distribution, whether shallow or deep, and optimize these prompts to drive the PVM to conduct downstream tasks. Compared to the prior art, our proposal (PRO-VPT) adaptively adjusts prompt distribution by treating it as an optimization objective and coupling the distribution optimization with prompt tuning.}
    \vspace{-0.23em}
    \label{fig:provpt_goal}
\end{figure}

Instead of updating all model parameters directly, VPT inserts lightweight learnable prompt tokens into the input space of Transformer blocks, with either a shallow distribution (inserted only in the first block) or a deep distribution (uniformly inserted across all blocks), while keeping the PVM frozen. This approach significantly reduces the number of parameters that need to be fine-tuned, thereby lowering the computational and memory overhead~\cite{han2024facing}. Follow-up research further enhances this pipeline by improving aspects such as prompting architecture~\cite{han2023e2vpt, shi2024dept, tang2025adept}, prompt initialization~\cite{wang2024revisiting}, and prompt propagation structure~\cite{zhou2024ivpt, xu2024progressive, das2023learning}.

Despite these advancements, most of the prior art still relies on a preset fixed prompt distribution, either shallow or deep, across various downstream tasks, as illustrated in Fig.~\ref{fig:provpt_goal}. However, recent findings reveal that the importance of each pre-trained block varies significantly across different tasks~\cite{shen2021partial, park2024layer, guo2019spottune, ko2022not, wanjiku2023dynamic}, underscoring the need to dynamically tailor the fine-tuning intensity for each block to improve task performance~\cite{lee2023surgical, tian2023trainable, gouk2021distancebased, ro2021autolr, kaplun2023less}. This implies that indiscriminately using a fixed prompt distribution may not fully exploit the potential of VPT, motivating us to adaptively adjust the distribution to better accommodate each task. 

In this paper, we naturally consider prompt distribution as one of the optimization objectives, which we refer to as adaptive distribution optimization (\textbf{ADO}). As the focus pivots towards the ADO problem, new questions arise. \textbf{First}, since the underlying nature of distribution optimization is still unexplored, the definition of ADO remains unclear. For instance, defining it as a one-shot process or an iterative one leads to a critical consideration, namely whether the effectiveness of distribution adjustments is influenced by ongoing prompt updates. This leads us to our first key question: \ding{172} \textit{How can we appropriately and formally define ADO?} (\S~\ref{sec:method:problem}) \textbf{Second}, once the definition of ADO is established, a more essential question emerges:  \ding{173} \textit{How can we achieve adaptive distribution derived from this definition?} (\S~\ref{sec:method:pr})

To address these questions, we first empirically analyze the ADO problem. Our findings confirm the significance of ADO and further reveal two key insights: the effectiveness of distribution adjustments is indeed impacted by prompt updates, and a nested relationship exists between ADO and VPT. 
Based on these insights, we propose \textbf{PRO-VPT} (iterative \underline{P}rompt \underline{R}el\underline{O}cation-based VPT), a novel iterative framework derived from a nested optimization definition. Specifically, building on this definition, we develop a prompt relocation (\textbf{PR}) strategy for ADO, which comprises two optimization steps: pruning and allocating. The pruning step first identifies and releases idle prompts from prompt-saturated blocks, while the subsequent allocation step determines the most prompt-needed blocks for redistributing these prompts. By iteratively performing PR and VPT, PRO-VPT can adaptively learn the optimal prompt distribution for each task in a nested optimization manner, thereby unlocking the full potential of VPT.
Our main contributions are fourfold.
\begin{itemize}
    \item We explore the ADO problem within VPT and reveal the nested relationship between ADO and VPT.
    \item Building upon our insights into prompt distribution, we propose the first ADO-VPT co-design framework, termed PRO-VPT, with a nested optimization definition.
    \item Derived from the definition, we present a PR strategy for ADO, which introduces an idleness score to prune underutilized prompts from prompt-saturated blocks and then allocates them to prompt-needed blocks using reinforcement learning (RL).
    \item Extensive experiments show that PRO-VPT exhibits superior performance, \textit{e.g.}, it attains state-of-the-art average accuracies of 78.0\% on VTAB-1k and 91.7\% on FGVC, with 1.6 pp and 2.0 pp advantages over the VPT baseline.
\end{itemize}

\vspace{-0.65em}
\section{Related Work}
\vspace{-0.35em}

\noindent \textbf{Visual Prompt Tuning.} 
VPT~\cite{jia2022visual}, one of the leading techniques in PEFT, draws inspiration from language-domain prompting works~\cite{liu2023pre, li-liang-2021-prefix, liu-etal-2022-p, hou2025capt} to fine-tune a small portion of prompts while keeping the pre-trained model frozen. Subsequent research has improved VPT in three main aspects: (1) Extending prompts to other Transformer components (\textit{e.g.}, key and value matrices or the patch embedding layer) to better suit the Transformer architecture~\cite{han2023e2vpt, shi2024dept, tang2025adept}; (2) Initializing prompts with prototypes of image tokens to accelerate convergence~\cite{wang2024revisiting}; (3) Refining the propagation structures of prompts to strengthen cross-layer interactions~\cite{zhou2024ivpt, xu2024progressive, das2023learning}. However, most existing methods depend on a fixed prompt distribution for all tasks, resulting in less effective prompting performance. In contrast, our approach adaptively adjusts the prompt distribution for each task, enabling prompt-based methods to achieve state-of-the-art performance.

\noindent \textbf{Layer-wise Fine-Tuning.}
Recent research has highlighted the importance of treating each pre-trained block differently during downstream fine-tuning, emphasizing the need for layer-wise adaptive strategies to calibrate the fine-tuning intensity for each block~\cite{tian2024fast, li2018delta, guo2019spottune, park2024layer, kaplun2023less}. For instance, \cite{lee2023surgical} first revealed that some pre-trained blocks may already be near-optimal for specific tasks, indicating that fine-tuning demands of each block vary significantly for different tasks. This work then proposes freezing these near-optimal blocks and selectively fine-tuning the remaining ones to improve performance. Similarly, \cite{shen2021partial, ro2021autolr} proposes to adaptively assign different learning rates to each block, tailoring the fine-tuning intensity at a layer level to achieve more fine-grained layer-wise adaptation. Inspired by these findings, we hypothesize that the prompting intensity for each block, \textit{i.e.}, the overall distribution of prompts, may also play a crucial role in task adaptation, leading us to focus on adaptively adjusting the prompt distribution within VPT.

\vspace{-0.5em}
\section{Methodology}
\label{sec:method}
\vspace{-0.35em}

\vspace{-0.1em}
\subsection{Notation} \label{sec:method:notation}
\vspace{-0.2em}

Visual prompt tuning (VPT) aims to learn lightweight tunable prompts $\mathbf{P}$ that are inserted into the vision Transformer (ViT) $f(\cdot)$ with a specific distribution $\mathcal{D}$. In particular, for a set of $N$ prompts $\mathbf{P}\!=\! \{ \mathbf{p}_k \}^N_{k=1}$ and their distributed blocks $\mathcal{D}\!=\! \{ d_k \}^N_{k=1}$ where $d_k\! \in\! \{ 0, 1, \ldots, L \}$ ($d_k\!=\!0$ for undistributed), the prompted ViT model is denoted as $f_{\mathbf{P}, \mathcal{D}}(\cdot)$. Given a training set $\mathcal{T}_{tr}$ of input-label pairs $(\mathbf{x}, y)$, the model is then trained by minimizing the task loss $\mathcal{L}(f_{\mathbf{P}, \mathcal{D}}(\mathbf{x}), y)$. For convenience in later sections, we define the loss difference as $\Delta \mathcal{L}(f_{1}, f_{2})\!=\! \mathcal{L}(f_{1}(\mathbf{x}), y) \!-\! \mathcal{L}(f_{2}(\mathbf{x}), y)$.

\vspace{-0.2em}
\subsection{Analysis and Formulation of ADO}  \label{sec:method:problem}
\vspace{-0.2em}

We first explore the underlying nature of ADO and its formal definition, in response to question \ding{172}. 

Intuitively, ADO can be defined in two distinct ways: \textbf{one-shot} (independently adjusting the distribution) or \textbf{iterative} (integrating with prompt tuning through iteratively adjusting the distribution). These definitions are distinct in whether the distribution is adjusted concurrently with prompt tuning. Specifically, an iterative process is more appropriate when distribution adjustments are affected by ongoing changes in prompts, as independent optimization may lead to suboptimal results by failing to account for these dynamics. Otherwise, one-shot adjustments may be sufficient.

\begin{figure}[t]
  \centering
  \includegraphics[scale=0.64]{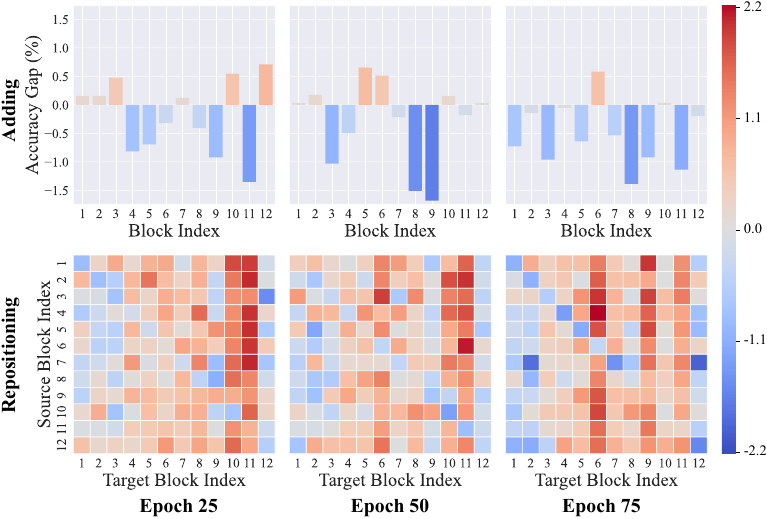}
  \vspace{-1.8em}
  \caption{
  \textbf{Performance gaps from distribution adjustments using prompts from epochs 25, 50, and 75.} Adjusting prompt distribution appropriately leads to enhanced performance; however, effective adjustments vary significantly across different epochs.}
  \vspace{-0.1em}
  \label{fig:ado_exp}
\end{figure}

To investigate which definition better matches the underlying nature, we experiment to examine how distribution adjustments behave when applied to prompts from different epochs. We use two intuitive adjustment strategies: incrementally \textit{adding} to a small initial prompt set and \textit{repositioning} existing prompts. As shown in Fig.~\ref{fig:ado_exp}, adding or repositioning prompts to specific blocks can effectively improve performance; however, the effective adjustments differ significantly when prompts are updated. Comprehensive results with detailed analysis are provided in Appendix~\ref{sec:app1}. In conclusion, we summarize our key findings as follows:

\noindent \textbf{Finding 1.} \textit{Properly adjusting the prompt distribution significantly enhances performance, confirming the necessity to reconsider its optimization.}

\noindent \textbf{Finding 2.} \textit{Distribution adjustments are sensitive not only to different blocks but also to the updated prompts themselves, highlighting the need for an iterative process that continuously adjusts the distribution during training.}

\noindent \textbf{Finding 3.} \textit{The effectiveness of distribution adjustments can only be properly evaluated after prompt tuning, indicating that the prompt tuning process should be nested within the distribution optimization process (see Appendix~\ref{sec:app1}).}

Building on these insights, we propose integrating ADO with VPT into an \textbf{iterative} optimization framework. Formally, we present the definition of this ADO-VPT co-design framework through the lens of \textbf{nested} optimization, where the objectives of ADO and VPT are expressed as:
\begin{align}
    \label{eq:distribution_learning}
    \mathcal{D}^\ast &= \mathop{\arg\min}\limits_{\mathcal{D}}~\mathbb{E}_{(\mathbf{x}, y) \in \mathcal{T}_{tr}}\big[\mathcal{L}(f_{\mathbf{P}^\ast, \mathcal{D}}(\mathbf{x}), y)\big], \\
    \label{eq:prompt_tuning}
    \mathbf{P}^\ast &= \mathop{\arg\min}\limits_{\mathbf{P}}~\mathbb{E}_{(\mathbf{x}, y) \in \mathcal{T}_{tr}}\big[\mathcal{L}(f_{\mathbf{P}, \mathcal{D}^\ast}(\mathbf{x}), y)\big].
\end{align}
This formulation clearly illustrates the intertwined relationship between visual prompts $\mathbf{P}$ and their distribution $\mathcal{D}$, with its framework presented in Fig.~\ref{fig:appendix13}. In this framework, prompt tuning is integrated into the ADO process, forming a closed-loop workflow of `distribution adjustment $\rightarrow$ prompt tuning $\rightarrow$ adjustment evaluation $\rightarrow$ new adjustment', thereby achieving a co-design process aligning with the observed iterative and nested nature.

\noindent \textbf{Remark.}
This nested optimization framework offers a universal approach capable of incorporating various distribution adjustment strategies, \textit{e.g.}, adding or repositioning. However, we empirically observe that adding-based adjustments often lead to unstable performance (see Appendix~\ref{sec:app2}). Therefore, in the following section, we focus on developing a prompt relocation (PR) strategy based on this formulation to effectively tackle the ADO problem.

\begin{figure*}[ht]
  \centering
  \includegraphics[scale=0.99]{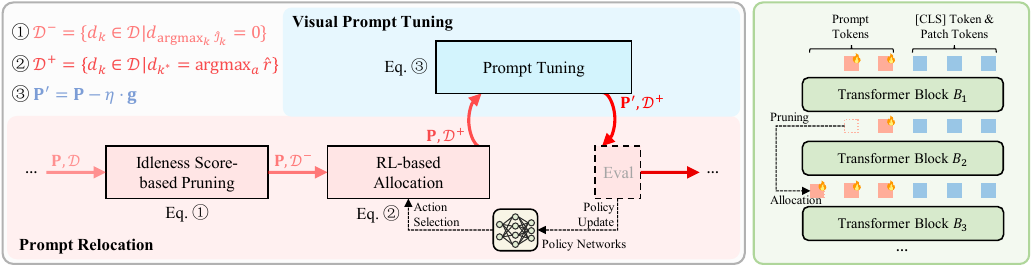}
  \vspace{-0.7em}
  \caption{\textbf{Overview of our proposed PRO-VPT}. 
  \textit{Left:} The streamlined workflow of PRO-VPT. \textit{Right:} Illustration of the PR process. 
  }
  \vspace{-1.3em}
  \label{fig:provpt_framework}
\end{figure*}

\vspace{-0.3em}
\subsection{Iterative Prompt Relocation-based VPT}  
\vspace{-0.2em}
\label{sec:method:pr}

Following the relocation objective defined in Eq.~\eqref{eq:distribution_learning}, we introduce the PR strategy derived from this definition and provide a detailed overview of the PRO-VPT framework, in response to question \ding{173}. 

Building on the aforementioned findings, the \textit{discrete} PR objective is further complicated by the \textit{non-stationary} nature of effective relocation (\textbf{Finding 2}) and the necessity to \textit{predict} the effectiveness of relocation operations (\textbf{Finding 3}). Nonetheless, these three properties make a RL framework~\cite{li2017deep, kaelbling1996reinforcement} a natural and well-suited approach for optimizing the PR objective. However, a naive RL implementation needs to model all possible relocation operations as actions, resulting in $L^2$ possible arrangements (\textit{e.g.}, 144 actions for ViT-B). Such a large action space necessitates numerous iterations to achieve convergence (\textit{cf.} \S~\ref{sec:exp:ablation}). 

To address this, we propose decomposing the PR process into two sequential steps: \textbf{pruning} and \textbf{allocating}. The pruning step first identifies and releases idle prompts for relocation, while the allocating step then employs RL to determine optimal blocks for these freed prompts. By focusing solely on allocatable blocks instead of all block arrangements, this approach effectively reduces the action space from $L^2$ to $L$. More importantly, this approach has an intuitive explanation: idle prompts typically exist within the prompt-saturated blocks, thus relocating them to prompt-needed blocks may enhance the effectiveness of prompt distribution.
Decomposed from the relocation objective in Eq.~\eqref{eq:distribution_learning}, the pruning and allocation steps\footnote{Here, for simplicity and efficiency, both the pruning and allocation steps are formalized to operate on a single prompt per iteration, as further elaborated in Appendix~\ref{sec:app3}.} can be formalized as:
\begin{align}
    \label{eq:pruning_goal}
    k^\ast = \ &\mathop{\arg\max}\limits_{k}~\mathbb{E}_{(\mathbf{x}, y) \in \mathcal{T}_{tr}} \big[\Delta \mathcal{L}(f_{\mathbf{P}, \mathcal{D}}, f_{\mathbf{P}, \mathcal{D}\mid d_k =0}) \big], \\
    \label{eq:allocating_goal}
    a^\ast = \ &\mathop{\arg\max}\limits_{a}~\mathbb{E}_{(\mathbf{x}, y) \in \mathcal{T}_{tr}} \big[\Delta \mathcal{L}(f_{\mathbf{P}, \mathcal{D}^-}, f_{\mathbf{P}^\prime, \mathcal{D}\mid d_{k^{\!\ast\!}} = a}) \big],
\end{align}
where $\mathcal{D}^{-}\!=\!\{ d_k \!\in\! \mathcal{D} \!\mid\! d_{k^{\!\ast\!}} \!=\! 0 \}$ and $\mathcal{D}^{+}\!=\!\{ d_k \!\in\! \mathcal{D} \!\mid\! d_{k^{\!\ast\!}} \!=\! a^\ast \}$ denote the prompt distributions after pruning and allocation with $a\!\in\![L]$ as the allocation block, and the allocation step uses the tuned prompts $\mathbf{P}^\prime$ for evaluation following Finding 3. These two equations are specifically designed to identify the $k^{*}$-th prompt suitable for pruning and to determine the optimal block $a^*$ for allocating.

Following this, we utilize Eqs.~\eqref{eq:pruning_goal} and \eqref{eq:allocating_goal} to approach the pruning and allocation stages, respectively, with the overall PRO-VPT framework illustrated in Fig.~\ref{fig:provpt_framework} and Algo.~\ref{alg:algo}:

\noindent \textbf{Idleness Score-based Pruning.} 
To measure whether a prompt is idle for its current block and suitable for relocation, we take Eq.~\eqref{eq:pruning_goal} as the idleness score:
\begin{equation}
    \mathcal{I}_{k} = \Delta \mathcal{L}(f_{\mathbf{P}, \mathcal{D}}, f_{\mathbf{P}, \mathcal{D}\mid d_k =0}).
\end{equation}
\textit{Intuitive Interpretation.}
If $\mathcal{I}_k \!=\! \Delta \mathcal{L} \!>\!0$, pruning the prompt $\mathbf{p}_k$ from its current block can enhance performance. This indicates that $\mathbf{p}_k$ is negative for its current block, which tends to be prompt-saturated and therefore suitable for its relocation to a more prompt-needed block. Conversely, $\mathcal{I}_k\!\leq\!0~\forall k$ suggests that no blocks are saturated and thus the prompt distribution is near-optimal. 

\noindent \textit{Pruning Implementation.}
Since computing $\mathcal{I}_k$ for each individual prompt $\mathbf{p}_k$ is significantly inefficient, we approximate $\mathcal{I}_{k}$ using its first-order Taylor expansion (see Appendix~\ref{sec:app4} for detailed derivation):
\begin{equation} \label{eq:idleness_score}
    \widehat{\mathcal{I}}_{k} \approx \mathbf{g}_{k}^T \mathbf{p}_{k},
\end{equation}
where $\mathbf{g}_{k}\!=\!\frac{\nabla \mathcal{L}}{\nabla \mathbf{p}_k}$ represents the elements of the gradient $\mathbf{g}$.

To this end, by calculating the estimated scores as in Eq.~\eqref{eq:idleness_score}, we can effectively identify and prune the idle prompt $\mathbf{p}_{k^{\!\ast\!}}$ with the maximum $\widehat{\mathcal{I}}_{k^{\!\ast\!}}$ from the prompt-saturated block (when $\max \widehat{\mathcal{I}}_k \!>\!0$), thereby fulfilling Eq.~\eqref{eq:pruning_goal} and forming the pruned distribution $\mathcal{D}^-$.

\noindent \textbf{RL-based Allocation.} 
After pruning an idle prompt, the following key step is to leverage RL to determine its optimal allocation to the prompt-needed block. Specifically, we take Eq.~\eqref{eq:allocating_goal} as the optimization objective for the RL problem, \textit{i.e.}, the allocation reward, as follows:
\begin{equation} \label{eq:expected_reward}
    r = \Delta \mathcal{L}(f_{\mathbf{P}, \mathcal{D}^-}, f_{\mathbf{P}^\prime, \mathcal{D}\mid d_{k^{\!\ast}\!=\!a}}).
\end{equation}
\textit{Intuitive Interpretation.}
The optimal allocation block $a^\ast$, which achieves the maximum $r^{\ast}\!=\!\max\Delta\mathcal{L}$, is identified as the block where assigning a prompt leads to the highest performance improvement, implying that this block is the most prompt-needed.

\noindent \textit{Allocation Implementation.}
We detail the RL-based allocation under a Markov decision process framework, where the main components are:

(1) State.
We leverage all available information to represent the state of the current prompt distribution. Specifically, the state incorporates the block-wise idleness scores, the current distribution, and the pruned prompt’s location, formally defined as $s \!=\! \{ \sum_{d_k\!=\!i} \widehat{\mathcal{I}}_{k} \}_{i\!=\!1}^L \!\bigcup\! \mathcal{D}^- \!\bigcup\! \text{\texttt{OneHot}}(k^\ast)$. After PR and VPT, the state transitions to $s^\prime$.

(2) Action.
The action determines which block the pruned prompt will be allocated to, represented as $a \in [L]$.

(3) Reward.
Although the allocation reward $r$ is defined in Eq.~\eqref{eq:expected_reward}, calculating the intermediate loss of $f_{\mathbf{P}, \mathcal{D}^-}$ is computationally inefficient. To address this, we approximate $r$ by incorporating the idleness score as:
\begin{equation}  \label{eq:reward}
    \hat{r} \approx \Delta \mathcal{L}(f_{\mathbf{P}, \mathcal{D}}, f_{\mathbf{P}^\prime, \mathcal{D}\mid d_{k^{\!\ast}\!=\!a}}) - \widehat{\mathcal{I}}_{k^{\!\ast}},
\end{equation}
where the idleness score $\widehat{\mathcal{I}}_{k^{\!\ast}}\!\approx\! \Delta \mathcal{L}(f_{\mathbf{P}, \mathcal{D}}, f_{\mathbf{P}, \mathcal{D}^-})$ provides an approximation of the intermediate loss of $f_{\mathbf{P}, \mathcal{D}^-}$.

To this end, by leveraging RL to maximize the estimated reward as in Eq.~\eqref{eq:reward}, we can efficiently determine the optimal allocation to the prompt-needed block $a^\ast$ even within the non-stationary environment, thereby fulfilling Eq.~\eqref{eq:allocating_goal} and forming the relocated distribution $\mathcal{D}^+$. Particularly, we accomplish this by employing proximal policy optimization (PPO)~\cite{schulman2017proximal}, a state-of-the-art RL algorithm.

\noindent \textbf{Remark.} 
Although the relocation process is decomposed into two-step sequential steps, the overall optimization objective, whether based on the expected objectives in Eqs.~\eqref{eq:pruning_goal} and \eqref{eq:allocating_goal} or the estimated objectives in Eqs.~\eqref{eq:idleness_score} and \eqref{eq:reward}, can be expressed as $\max_{k, a} \mathbb{E}_{(\mathbf{x}, y) \in \mathcal{T}_{tr}} [\Delta \mathcal{L}(f_{\mathbf{P}, \mathcal{D}}, f_{\mathbf{P}^\prime, \mathcal{D}\mid d_{k}\!=\!a})]$. This formulation remains \textit{equivalent} to the original PR objective (\textit{i.e.}, the ADO objective) defined in Eq.~\eqref{eq:distribution_learning}. Consequently, leveraging the nested optimization formulation, the proposed PR strategy can effectively address the ADO problem and adaptively construct the optimal distribution.

\vspace{-0.25em}
\subsection{Discussion} \label{sec:discussion}
\vspace{-0.15em}

We note that some prior works have \textit{indirectly} achieved adaptive distribution~\cite{han2023e2vpt, kim2024we, zhang2024neural}. Specifically, pruning-based techniques~\cite{han2023e2vpt, kim2024we} introduce mask variables to iteratively prune redundant prompts from the original distribution, and NOAH~\cite{zhang2024neural} uses an evolutionary algorithm to search for an optimal sub-distribution within the original distribution in a one-shot manner. PRO-VPT is distinct and advantageous compared to these methods as:

\noindent \textbf{Distinct Goals.} 
These methods aim to enhance performance through the lottery ticket hypothesis~\cite{frankle2018the, malach2020proving, savarese2020winning}, focusing on searching for a more effective sub-distribution within a given hyper-distribution by circumventing negative prompts and retaining essential ones. However, they are limited to pruning prompts from prompt-saturated blocks and fail to tackle prompt-deficient blocks (\textit{cf.} \S~\ref{sec:exp:understanding}). In contrast, our approach actively learns an optimal distribution of all prompts, adaptively relocating them from prompt-saturated blocks to prompt-needed blocks through our PR strategy, thereby maximizing the effectiveness of the entire prompt distribution. This fundamental distinction sets our approach apart from the methods discussed above.

\noindent \textbf{Efficient.}
While effective, we argue that these approaches typically incur considerable unnecessary training costs to identify an optimal sub-distribution, as numerous trained prompts are pruned and left unused (\textit{cf.} \S~\ref{sec:exp:ablation}). This fundamentally contradicts the goal of parameter efficiency. Conversely, our work directly adjusts the distribution of the entire prompt set through prompt relocation, thereby minimizing extra costs and improving efficiency.

\noindent \textbf{Effective.}
These methods straightforwardly apply one-shot or iterative processes without considering the underlying distribution dynamics. For instance, NOAH~\cite{zhang2024neural} adopts a one-shot approach post-training, leading to suboptimal results (\textit{cf.} \S~\ref{sec:exp:main_results}). Conversely, our work conducts empirical analyses and accordingly proposes a \textit{nested optimization-based} method, enabling prompt-based methods to reach state-of-the-art performance.

\vspace{-0.5em}
\section{Experiments}
\label{sec:exp}
\vspace{-0.2em}

\begin{table*}[t]
    \caption{\textbf{Detailed results on the VTAB-1k datasets.} Performance results are reported as the highest of ImageNet normalization ($\circ$) or Inception normalization ($\bullet$), presented in \% after a complete training schedule with ViT-B/16 supervised pre-trained on ImageNet-21k. The best results of prompt-based methods and other PEFT approaches are highlighted in \textbf{bold}. \Lightning: Early-stopping based on the test set. $\dagger$: Lack of complete code or hyperparameter configurations for the method, hence results are reported as presented in the original paper. $^1$Average across the average accuracies of the VTAB-1k groups, following previous work.}
    \vspace{-0.7em}
    \label{tab:vtab1k_results}
    \scriptsize
    \renewcommand*{\arraystretch}{0.95}
    \begin{tabularx}{\linewidth}{@{}+X-c@{\enspace\quad}-c@{\enspace}-c@{\enspace}-c@{\enspace}-c@{\enspace}-c@{\enspace}-c@{\enspace}-c@{\quad}-c@{\quad\enspace}-c@{\enspace}-c@{\enspace}-c@{\enspace}-c@{\quad}-c@{\quad\enspace}-c@{\enspace}-c@{\enspace}-c@{\enspace}-c@{\enspace}-c@{\enspace}-c@{\enspace}-c@{\enspace}-c@{\quad}-c@{\quad\enspace}-c@{}}
    \toprule
        & & \multicolumn{8}{c@{\enspace\quad}}{\textbf{Natural}} & \multicolumn{5}{c@{\enspace\quad}}{\textbf{Specialized}} & \multicolumn{9}{c@{\enspace\quad}}{\textbf{Structured}} \\
        \cmidrule(lr{2em}){3-10} \cmidrule(lr{2em}){11-15} \cmidrule(lr{2em}){16-24} \\ \addlinespace[-1em]
        & \rotatebox{90}{Param (M)} & \rotatebox{90}{Cifar100} & \rotatebox{90}{Caltech101} & \rotatebox{90}{DTD} & \rotatebox{90}{Flower102} & \rotatebox{90}{Pets} & \rotatebox{90}{SVHN}  & \rotatebox{90}{Sun397} & \rotatebox{90}{Group Avg.} & \rotatebox{90}{Camelyon} & \rotatebox{90}{EuroSAT}   & \rotatebox{90}{Resisc45}  & \rotatebox{90}{Retinopathy} & \rotatebox{90}{Group Avg.} & \rotatebox{90}{Clevr-Count} & \rotatebox{90}{Clevr-Dist.}  & \rotatebox{90}{DMLab} & \rotatebox{90}{KITTI-Dist.}  & \rotatebox{90}{dSpr-Loc.} & \rotatebox{90}{dSpr-Ori.}   & \rotatebox{90}{sNORB-Azi.}  & \rotatebox{90}{sNORB-Ele.} & \rotatebox{90}{Group Avg.} & \rotatebox{90}{Global Avg.$^1$}   \\
    \midrule
        Full $\bullet$ & 85.8 & 73.2 & 92.6 & 70.4 & 97.9 & 86.2 & 90.6 & 39.6 & 78.6 & 87.1 & 96.6 & 87.5 & 74.0 & 86.3 & 66.6 & 61.0 & 49.8 & 79.7 & 82.6 & 51.9 & 33.5 & 37.0 & 57.8 & 74.2 \\
        Linear $\bullet$ & 0.04 & 78.1 & 88.1 & 69.0 & 99.1 & 90.0 & 36.0 & 56.9 & 73.9 & 79.8 & 90.7 & 73.7 & 73.7 & 79.5 & 32.4 & 30.5 & 35.9 & 61.9 & 11.2 & 26.2 & 14.3 & 24.5 & 29.6 & 61.0 \\
    \midrule
        LoRA $\bullet$ \cite{hu2022lora} & 0.29 & 83.0 & 91.7 & 71.6 & 99.2 & 90.9 & 83.8 & 56.7 & 82.4 & 86.2 & 95.7 & 83.5 & 71.9 & 84.3 & 77.7 & 62.3 & 49.0 & 80.2 & 82.2 & 51.7 & 31.0 & 47.0 & 60.1 & 75.6 \\
        $\text{FacT-TK}_{8}$ $\bullet$ \cite{jie2023fact} & \textbf{0.05} & 74.9 & 92.7 & 73.7 & 99.1 & 91.3 & 85.5 & \textbf{57.7} & 82.1 & 86.8 & 94.9 & 84.1 & 70.9 & 84.2 & 81.9 & 64.1 & 49.2 & 77.2 & 83.8 & 53.1 & 28.2 & 44.7 & 60.3 & 75.5 \\
        FacT-TK$_{\leq 32}$ $\bullet$ \cite{jie2023fact} & 0.10 & 74.6 & 93.7 & 73.6 & 99.3 & 90.6 & 88.7 & 57.5 & 82.6 & \textbf{87.6} & 95.4 & 85.5 & 70.4 & 84.7 & \textbf{84.3} & 62.6 & 51.9 & 79.2 & 85.5 & 52.0 & 36.4 & 46.6 & 62.3 & 76.5 \\
        Consolidator $\dagger$ \cite{hao2023consolidator} & 0.30 & 74.2 & 90.9 & 73.9 & \textbf{99.4} & 91.6 & \textbf{91.5} & 55.5 & 82.4 & 86.9 & 95.7 & 86.6 & \textbf{75.9} & 86.3 & 81.2 & \textbf{68.2} & 51.6 & \textbf{83.5} & 79.8 & 52.3 & 31.9 & 38.5 & 60.9 & 76.5 \\
        SSF \Lightning$\circ$ \cite{lian2022scaling} & 0.24 & 69.0 & 92.6 & \textbf{75.1} & \textbf{99.4} & \textbf{91.8} & 90.2 & 52.9 & 81.6 & 87.4 & 95.9 & \textbf{87.4} & 75.5 & \textbf{86.6} & 75.9 & 62.3 & 53.3 & 80.6 & 77.3 & 54.9 & 29.5 & 37.9 & 59.0 & 75.7 \\
        SPT-Adapter \Lightning$\circ$ \cite{he2023sensitivity} & 0.23 & 72.9 & 93.2 & 72.5 & 99.3 & 91.4 & 84.6 & 55.2 & 81.3 & 85.3 & 96.0 & 84.3 & 75.5 & 85.3 & 82.2 & 68.0 & 49.3 & 80.0 & 82.4 & 51.9 & 31.7 & 41.2 & 60.8 & 75.8 \\
        SPT-Adapter \Lightning$\circ$ \cite{he2023sensitivity} & 0.43 & 72.9 & 93.2 & 72.5 & 99.3 & 91.4 & 88.8 & 55.8 & 82.0 & 86.2 & 96.1 & 85.5 & 75.5 & 85.8 & 83.0 & 68.0 & 51.9 & 81.2 & 82.4 & 51.9 & 31.7 & 41.2 & 61.4 & 76.4 \\
        Adapter+$_{r=16}$ $\bullet$ \cite{steitz2024adapters} & 0.35 & \textbf{83.7} & \textbf{94.2} & 71.5 & 99.3 & 90.6 & 88.2 & 55.8 & \textbf{83.3} & 87.5 & \textbf{97.0} & \textbf{87.4} & 72.9 & 86.2 & 82.9 & 60.9 & \textbf{53.7} & 80.8 & \textbf{88.4} & \textbf{55.2} & \textbf{37.3} & \textbf{46.9} & \textbf{63.3} & \textbf{77.6} \\

    \midrule
    \textcolor{\tblgray}{Prompt-based Methods:} & & & & & & & & & & & & & & & & & & & & & & & & \\
        VPT-Deep $\bullet$ \cite{jia2022visual} & 0.60 & 83.0 & 93.0 & 71.2 & 99.0 & 91.3 & 84.1 & 56.0 & 82.5 & 84.9 & 96.6 & 82.5 & 74.5 & 84.6 & 77.5 & 58.7 & 49.7 & 79.6 & 86.2 & 56.1 & 37.9 & 50.7 & 62.1 & 76.4 \\
        NOAH \Lightning$\circbullet$ \cite{zhang2024neural} & 0.43 & 69.6 & 92.7 & 70.2 & 99.1 & 90.4 & 86.1 & 53.7 & 80.2 & 84.4 & 95.4 & 83.9 & 75.8 & 84.9 & \textbf{82.8} & \textbf{68.9} & 49.9 & \textbf{81.7} & 81.8 & 48.3 & 32.8 & 44.2 & 61.3 & 75.5 \\
        SPT-Deep $\dagger$ \cite{wang2024revisiting}  & \textbf{0.22} & 79.3 & 92.6 & \textbf{73.2} & \textbf{99.5} & 91.0 & 89.1 & 51.2 & 82.3 & 85.4 & \textbf{96.8} & 84.9 & 74.8 & 85.5 & 70.3 & 64.8 & \textbf{54.2} & 75.2 & 79.3 & 49.5 & 36.5 & 41.5 & 58.9 & 75.6 \\
        iVPT $\dagger$ \cite{zhou2024ivpt} & 0.60 & 82.7 & \textbf{94.2} & 72.0 & 99.1 & \textbf{91.8} & 88.1 & 56.6 & 83.5 & \textbf{87.7} & 96.1 & \textbf{87.1} & \textbf{77.6} & \textbf{87.1} & 77.1 & 62.6 & 49.4 & 80.6 & 82.1 & 55.3 & 31.8 & 47.6 & 60.8 & 77.1 \\
        PRO-VPT (\textbf{ours}) & 0.61 & \textbf{84.5} & 94.1 & \textbf{73.2} & 99.4 & \textbf{91.8} & \textbf{88.2} & \textbf{57.2} & \textbf{84.1} & \textbf{87.7} & \textbf{96.8} & 86.6 & 75.5 & 86.7 & 78.8 & 61.0 & 50.6 & 81.3 & \textbf{86.7} & \textbf{56.4} & \textbf{38.1} & \textbf{51.7} & \textbf{63.1} & \textbf{78.0} \\
    \bottomrule
    \end{tabularx}
    \vspace{-1.6em}
\end{table*}

\vspace{-0.1em}
\subsection{Experiment Setups}
\vspace{-0.2em}
\noindent \textbf{Datasets.}
We evaluated our proposed PRO-VPT on a series of datasets grouped into three categories: (1) \textbf{VTAB-1k.} The visual task adaptation benchmark (VTAB)~\cite{zhai2019large} encompasses 19 classification tasks categorized into three group: \textit{Natural}, \textit{Specialized}, and \textit{Structured}, aiming to assess adaptation capabilities under low-shot training conditions. (2) \textbf{FGVC.} Fine-grained visual classification (FGVC) includes 5 classification tasks: CUB-200-2011~\cite{wah2011caltech}, NABirds~\cite{van2015building}, Oxford Flowers~\cite{nilsback2008automated}, Stanford Dogs~\cite{dataset2011novel}, and Stanford Cars~\cite{gebru2017fine}, aiming to assess task adaptation in large-scale, fine-grained settings. (3) \textbf{COCO and ADE20k.} These widely-used benchmarks for detection and segmentation are employed to assess the generalizability of our method across diverse downstream visual tasks.

\noindent \textbf{Implementation Details.}
We mainly use ViT-B/16~\cite{dosovitskiy2020vit} supervised pre-trained on ImageNet-21k~\cite{russakovsky2015imagenet} as the initialization. We employ the SGD optimizer with a batch size of 64 and fine-tune models for 100 epochs. The number of prompt tokens per task in our method is set in accordance with~\cite{jia2022visual}. Following~\cite{steitz2024adapters}, we evaluate our method using Inception normalization instead of ImageNet normalization, and for fair comparisons, we present the results of other counterparts using both normalization techniques. Notably, unlike most approaches~\cite{zhao2024sct, fu2024dtl, jia2022visual, lian2022scaling}, our method employs a fixed learning rate instead of using a learning rate schedule, as we necessitate ensuring timely rewinding after prompt relocation~\cite{renda2020comparing}.
More details about the hyper-parameters are provided in Appendices~\ref{sec:app5}.

\vspace{-0.3em}
\subsection{Main Results}  \label{sec:exp:main_results}
\vspace{-0.1em}

\noindent \textbf{VTAB-1k. } 
We evaluate the proposed PRO-VPT on VTAB-1k and compare it with other methods. We provide results for \textit{full} fine-tuning and \textit{linear} probing as baselines of traditional fine-tuning. As competing advanced adapter- or reparameterization-based methods, we include SSF~\cite{lian2022scaling}, SPT-Adapter~\cite{he2023sensitivity}, Adapter+~\cite{steitz2024adapters}, LoRA~\cite{hu2022lora}, FacT-TK~\cite{jie2023fact}, and Consolidator~\cite{hao2023consolidator}. Furthermore, we benchmark against state-of-the-art prompt-based methods including VPT~\cite{jia2022visual}, NOAH~\cite{zhang2024neural}, SPT~\cite{wang2024revisiting}, and iVPT~\cite{zhou2024ivpt}. 

Tab.~\ref{tab:vtab1k_results} presents results on VTAB-1k with the best-performing normalization techniques (\textit{i.e.}, ImageNet or Inception) for each method (see Appendix~\ref{sec:app6} for complete results). Among all prompt-based methods, the previous best performance of 77.1\% average accuracy achieved by iVPT falls short of the advanced adapter-based method Adapter+, which reaches 77.6\%. Nevertheless, with the introduction of the optimization-based adaptive distribution, the proposed PRO-VPT sets a new state-of-the-art performance among all evaluated methods with an average accuracy of 78.0\%, highlighting the significant potential of prompt-based approaches. Additionally, compared to the one-shot optimization-based NOAH, our nested optimization-based PRO-VPT achieves an average accuracy gain of 2.5 pp, validating the effectiveness of our proposed nested optimization framework.

\begin{figure}[t]
  \centering
  \hspace{-0.75em} \includegraphics[scale=0.23]{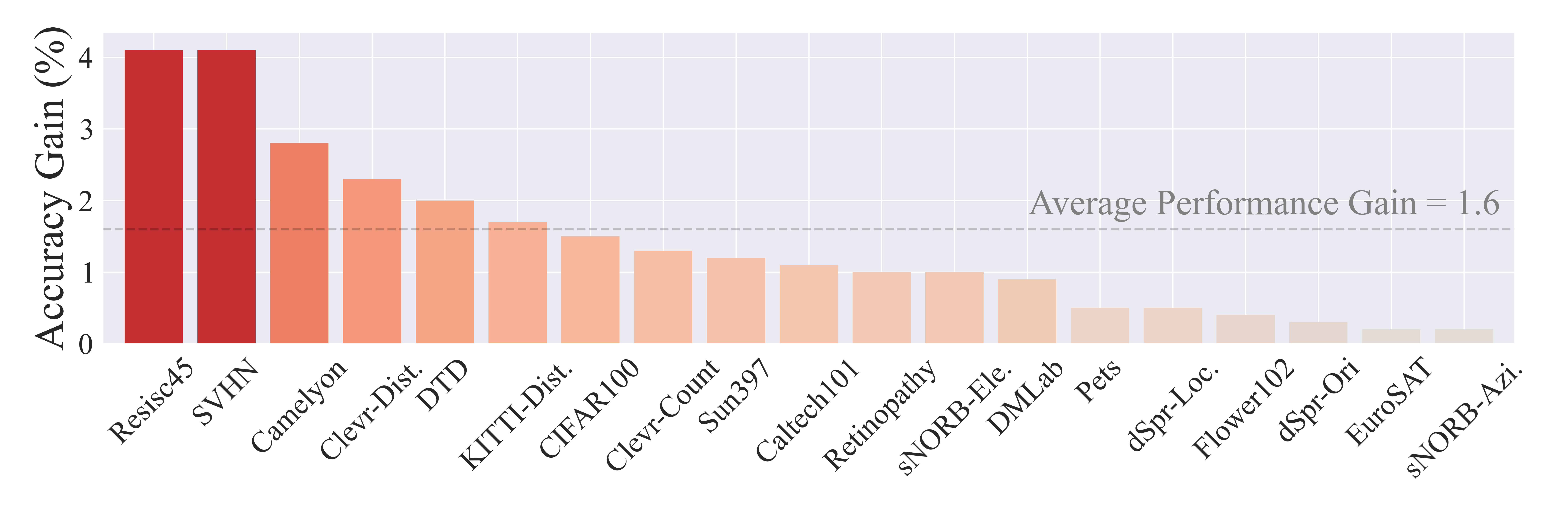}
  \vspace{-2.2em}
  \caption{\textbf{Performance gains} achieved by VPT w/ ADO (PRO-VPT) compared to VPT w/o ADO (VPT-Deep). PRO-VPT consistently outperforms VPT-Deep.}
  \label{fig:performance_gain}
\end{figure}

Furthermore, Fig.~\ref{fig:performance_gain} illustrates the performance differences between VPT w/ ADO (PRO-VPT) and w/o ADO (VPT-Deep) on the VTAB-1k datasets. The accuracies of PRO-VPT are consistently higher than those of VPT-Deep, yielding an average accuracy improvement of 1.6 pp. This emphasizes the significance of optimizing the prompt distribution to unlock the full potential of VPT.

\begin{table}[t]
    \caption{\textbf{Detailed results on the FGVC datasets.} Performance results are reported as the highest of ImageNet normalization ($\circ$) or Inception normalization ($\bullet$), presented in \% after a complete training schedule with ViT-B/16 supervised pre-trained on ImageNet-21k. The best results of prompt-based methods and other PEFT approaches are highlighted in \textbf{bold}. $\dagger$: Same as Tab.~\ref{tab:vtab1k_results}.}
    \vspace{-0.7em}
    \label{tab:fgvc_results}
    \footnotesize
    \renewcommand*{\arraystretch}{0.95}
    \begin{tabularx}{\columnwidth}{@{}+X-c@{\enspace\quad}-c@{\enspace}-c@{\enspace}-c@{\enspace}-c@{\enspace}-c@{\enspace\quad}-c@{}}
    \toprule
        & \rotatebox{90}{Param (M)} & \rotatebox{90}{CUB200} & \rotatebox{90}{NABirds} & \rotatebox{90}{Oxford Flowers} & \rotatebox{90}{Stanford Dogs }& \rotatebox{90}{Stanford Cars} & \rotatebox{90}{Global Avg.} \\
    \midrule
    Full $\bullet$ & 86.0 & 88.0 & 81.5 & 99.2 & 85.6 & 90.6 & 89.0 \\
    Linear $\bullet$ & 0.18  & 88.9 & 81.8 & 99.5 & 92.6 & 52.8 & 83.1 \\
    \midrule
    SSF $\circ$ \cite{lian2022scaling} & 0.39 & 89.5 & \textbf{85.7} & 99.6 & 89.6 & \textbf{89.2} & 90.7 \\
    SPT-Adapter $\dagger$ \cite{he2023sensitivity} & 0.40 & 89.1 & 83.3 & 99.2 & 91.1 & 86.2 & 89.8 \\
    SPT-LoRA $\dagger$ \cite{he2023sensitivity} & 0.52 & 88.6 & 83.4 & 99.5 & 91.4 & 87.3 & 90.1 \\
    Adapter+ $\bullet$ \cite{steitz2024adapters} & \textbf{0.34} & \textbf{90.4} & 85.0 & \textbf{99.7} & \textbf{92.6} & 89.1 & \textbf{91.4} \\
    \midrule
    \textcolor{\tblgray}{Prompt-based Methods:} & & & & & & & \\
    VPT-Deep $\bullet$ \cite{jia2022visual} & 0.85 & 90.1 & 83.3 & 99.6 & 90.3 & 85.0 & 89.7 \\
    SPT-Deep $\dagger$ \cite{wang2024revisiting} & \textbf{0.36} & \textbf{90.6} & \textbf{87.6} & \textbf{99.8} & 89.8 & 89.2 & 91.4 \\
    iVPT $\dagger$ \cite{zhou2024ivpt} & 0.41 & 89.1 & 84.5 & 99.5 & 90.8 & 85.6 & 89.9 \\
    PRO-VPT (\textbf{ours}) & 0.86 & \textbf{90.6} & 86.7 & 99.7 & \textbf{91.8} & \textbf{89.6} & \textbf{91.7} \\
    \bottomrule
    \end{tabularx}
\end{table}

\noindent \textbf{FGVC.} 
Next, we evaluate PRO-VPT on the FGVC benchmark. Among the aforementioned contenders, only a portion have conducted evaluations on FGVC and released the code and configurations.

Tab.~\ref{tab:fgvc_results} presents the FGVC results with the best-performing normalization (see Appendix~\ref{sec:app6} for complete results). PRO-VPT achieves the best average accuracy of 91.7\% over all five FGVC datasets, surpassing the second-best performance of 91.4\% achieved by Adapter+ and SPT-Deep. This demonstrates that PRO-VPT also delivers state-of-the-art results for task adaptation when training data is large-scale.

\begin{table}[t]
    \caption{\textbf{Ablation study on various alternative ADO strategies.} Results are presented using ViT-B/16 on two instances: \textit{Natural} SVHN and \textit{Specialized} Resisc45. The best and second-best results are highlighted in \textbf{bold} and \underline{underline}. $^\star$Our proposed PR strategy.}
    \vspace{-0.8em}
    \label{tab:ablation}
    \footnotesize
    \renewcommand*{\arraystretch}{1.0}
    \begin{tabularx}{\linewidth}{@{}Xcccc@{}}
    \toprule
    \multirow{2}{*}{\shortstack{\vspace{0.2em}\\\hspace{-1em}Alternatives for\\the ADO Problem}} & \multicolumn{2}{c}{SVHN} & \multicolumn{2}{c}{Resisc45} \\
    \cmidrule(lr){2-3} \cmidrule(lr){4-5} \\ \addlinespace[-1em]
    & Param (M) & Acc (\%) & Param (M) & Acc (\%) \\
    \midrule
    \textcolor{\tblgray}{VPT Baseline} & \textcolor{\tblgray}{0.468} & \textcolor{\tblgray}{84.10} & \textcolor{\tblgray}{0.127} & \textcolor{\tblgray}{82.50} \\
    (\texttt{A}) Pruning & \textbf{0.468} & 83.76 & \textbf{0.127} & 83.81 \\
    (\texttt{B}) Pruning & 2.543 & 84.17 & 0.541 & 84.86 \\
    (\texttt{C}) \textit{Naive RL} & \underline{0.482} & 83.51 & \underline{0.140} & 83.02 \\
    (\texttt{D}) \textit{Prn1} \texttt{\&} \textit{Alc1} & \textbf{0.468} & 84.88 & \textbf{0.127} & 84.06 \\
    (\texttt{E}) \textit{Prn1} \texttt{\&} \textit{Alc2} & \underline{0.482} & \underline{86.99} & \underline{0.140} & \underline{85.68} \\
    (\texttt{F}) \textit{Prn2} \texttt{\&} \textit{Alc1} & \textbf{0.468} & 85.82 & \textbf{0.127} & 84.76 \\
    (\texttt{G}) \textit{Prn2} \texttt{\&} \textit{Alc2}$^\star$ & \underline{0.482} & \textbf{88.19} & \underline{0.140} & \textbf{86.56} \\
    \bottomrule
    \end{tabularx}
\end{table}

\vspace{-0.3em}
\subsection{Ablation Study} \label{sec:exp:ablation}
\vspace{-0.4em}

To comprehensively assess the advantages of our proposed PR strategy, we compare it with the following alternative strategies, as shown in Tab.~\ref{tab:ablation}:

Strategies (\texttt{A}), (\texttt{B}) represent alternatives for extracting a sub-distribution from a hyper-distribution (see \S~\ref{sec:discussion}). We utilize the pruning-based technique as in~\cite{han2023e2vpt}, applying pruning percentages ranging from 10\% to 90\% with 10 pp intervals. (\texttt{B}) prunes from a huge hyper-distribution to obtain a sub-distribution with the same \textit{final} parameter count as the VPT baseline, and (\texttt{A}) prunes from a smaller hyper-distribution to match the number of \textit{trained} parameters in the VPT baseline.

Strategies (\texttt{C})-(\texttt{G}) represent alternatives of prompt relocating, each incorporating different components. Particularly, (\texttt{C}) frames the PR process as a \textit{naive RL} problem (see \S~\ref{sec:method:pr}) and utilizes the PPO algorithm. (\texttt{D})-(\texttt{G}) are various combinations of the following components:
{ \small
\begin{description}
    \vspace{-0.3em}
    \item[\textit{Prn1}.] Randomly pruning prompts.
    \vspace{-0.3em}
    \item[\textit{Prn2}.] Pruning underutilized prompts based on the idleness score.
    \vspace{-0.3em}
    \item[\textit{Alc1}.] Modeling the allocation task as a multi-armed bandit problem, which is well-suited for stationary discrete decision-making, and then adopting the widely used Thompson sampling for prompt allocation.
    \vspace{-0.3em}
    \item[\textit{Alc2}.] Modeling the allocation task as a vanilla RL problem, which is more suited to non-stationary decision-making, and then using the popular PPO algorithm for prompt allocation.
    \vspace{-0.3em}
\end{description}}

In summary, the key observations are as follows:

\noindent \textbf{Observation 1.} 
\textit{Relocation-based methods are more efficient and effective compared to pruning-based methods. } 
Pruning strategies (\texttt{A}) and (\texttt{B}) typically underperform relocation strategies (\texttt{D})-(\texttt{G}), indicating that pruning is less effective in adaptive distribution. Specific to pruning, while (\texttt{B}) outperforms (\texttt{A}), it incurs high storage costs (\textit{e.g.}, 2.075M trained but unused parameters in SVHN) and still lags behind our proposed PR strategy. In contrast, RL-based relocation strategies (\texttt{E}) and (\texttt{G}) require only 0.014M extra parameters for PPO's policy networks, which is relatively negligible.

\noindent \textbf{Observation 2.} 
\textit{The decomposition for RL is remarkably effective.} 
Strategy (\texttt{C}) significantly underperforms the two-step PR strategies (\texttt{D})-(\texttt{G}), underscoring the benefits of decomposing the PR process for simplifying decision-making.

\noindent \textbf{Observation 3.} 
\textit{Both the pruning and allocation components in PRO-VPT exhibit significantly effective, especially RL-based allocation.} 
Comparing strategies (\texttt{D}) vs. (\texttt{F}) and (\texttt{E}) vs. (\texttt{G}), \textit{Prn2} indicates superior performance improvements (averaging 1.07 pp for SVHN and 0.79 pp for Resisc45), highlighting the effectiveness of the proposed idleness score. Moreover, strategies (\texttt{G}) and (\texttt{E}) achieve the best results, with \textit{Alc2} delivering remarkable gains over \textit{Alc1} (2.24 pp and 1.71 pp), highlighting the advantage of framing the allocation task as a non-stationary RL problem.

\begin{table}[t]
    \centering
    \caption{\textbf{Generalizability study on distinct backbones.} Results are presented in \% on two instances: VTAB-1k \textit{Natural} Cifar100 and \textit{Structured} DMLab.}
    \vspace{-0.8em}
    \label{tab:backbone}
    \footnotesize
    \renewcommand*{\arraystretch}{1.0}
    \begin{tabular}{llll}
    \toprule
    Backbone & & Cifar100 & DMLab \\
    \midrule
    \multirow{2}{*}{ViT-B/16~~~~~~~} & VPT-Deep~ & 83.00 & 49.70 \\
    & \cellcolor{gray!20}PRO-VPT~ & \cellcolor{gray!20}84.49 \textcolor{tblblue}{\text{1.49} $\uparrow$} & \cellcolor{gray!20}50.64 \textcolor{tblblue}{\text{0.94} $\uparrow$} \\
    \midrule
    \multirow{2}{*}{ViT-L/16~~~~~~~} & VPT-Deep & 85.82 & 45.98 \\
    & \cellcolor{gray!20}PRO-VPT& \cellcolor{gray!20}87.07 \textcolor{tblblue}{\text{1.25} $\uparrow$} & \cellcolor{gray!20}46.57 \textcolor{tblblue}{\text{0.59} $\uparrow$} \\
    \midrule
    \multirow{2}{*}{ViT-H/14~~~~~~~} & VPT-Deep & 78.49 & 44.12\\
    & \cellcolor{gray!20}PRO-VPT& \cellcolor{gray!20}79.73 \textcolor{tblblue}{\text{1.24} $\uparrow$} & \cellcolor{gray!20}45.17 \textcolor{tblblue}{\text{1.05} $\uparrow$} \\
    \midrule
    \multirow{2}{*}{Swin-B~~~~~~~} & VPT-Deep & 81.33 & 49.86\\
    & \cellcolor{gray!20}PRO-VPT& \cellcolor{gray!20}82.67 \textcolor{tblblue}{\text{1.34} $\uparrow$} & \cellcolor{gray!20}50.78 \textcolor{tblblue}{\text{0.92} $\uparrow$} \\
    \bottomrule
    \end{tabular}
    \vspace{-0.5em}
\end{table}

\begin{table}[t]
    \centering
    \caption{\textbf{Generalizability study on different pre-training strategies.} Results are presented in \% on two instances: VTAB-1k \textit{Natural} Cifar100 and \textit{Structured} DMLab.}
    \vspace{-0.8em}
    \label{tab:pretrain}
    \footnotesize
    \renewcommand*{\arraystretch}{1.0}
    \begin{tabular}{llll}
    \toprule
    \multicolumn{2}{l}{Pre-Training Strategy} & Cifar100 & DMLab \\
    \midrule
    \multirow{2}{*}{Supervised~~~~~~} & VPT-Deep~ & 83.00 & 49.70 \\
    & \cellcolor{gray!20}PRO-VPT~ & \cellcolor{gray!20}84.49 \textcolor{tblblue}{\text{1.49} $\uparrow$} & \cellcolor{gray!20}50.64 \textcolor{tblblue}{\text{0.94} $\uparrow$} \\
    \midrule
    \multicolumn{2}{l}{\textcolor{\tblgray}{Self-Supervised:}} & & \\
    \multirow{2}{*}{MAE~~~~~~} & VPT-Deep & 32.65 & 43.58 \\
    & \cellcolor{gray!20}PRO-VPT& \cellcolor{gray!20}34.36 \textcolor{tblblue}{\text{1.71} $\uparrow$} & \cellcolor{gray!20}46.07 \textcolor{tblblue}{\text{2.49} $\uparrow$} \\
    \midrule
    \multirow{2}{*}{MoCo-v3~~~~~~} & VPT-Deep  & 72.48 & 46.56 \\
    & \cellcolor{gray!20}PRO-VPT& \cellcolor{gray!20}73.62 \textcolor{tblblue}{\text{1.14} $\uparrow$} & \cellcolor{gray!20}47.63 \textcolor{tblblue}{\text{1.07} $\uparrow$} \\
    \bottomrule
    \end{tabular}
\end{table}

\subsection{Generalizability Analysis} \label{sec:exp:generalize}
\vspace{-0.3em}
We evaluate the generalizability of PRO-VPT across different backbones and pre-training strategies, as well as in detection and segmentation tasks.

\noindent \textbf{Backbone.}
We experiment with three different scales of the ViT backbone: ViT-Base (ViT-B/16), ViT-Large (ViT-L/16), and ViT-Huge (ViT-H/14)~\cite{dosovitskiy2020vit}, along with the hierarchical Swin Transformer (Swin-B)~\cite{liu2021swin}. All backbones are supervised pre-trained on ImageNet-21k~\cite{russakovsky2015imagenet}. As shown in Tab.~\ref{tab:backbone}, PRO-VPT delivers higher accuracy compared to VPT with various scales and architectures.

\noindent \textbf{Pre-Training Strategy.}
In addition to the supervised pre-training strategy, we experiment with two self-supervised strategies: the masked image modeling method (MAE)~\cite{he2022masked} and the contrastive self-supervised method (MoCo-v3)~\cite{chen2021empirical}. As illustrated in Tab.~\ref{tab:pretrain}, PRO-VPT consistently outperforms VPT regardless of the pre-training strategy, highlighting its strong generalizability.

\noindent \textbf{Detection and Segmentation.} 
We also experiment on detection and segmentation tasks in Appendix~\ref{sec:app6}, where PRO-VPT still maintains advantages over VPT.

\begin{figure*}[!t]
  \centering
  \hspace{-0.9em} \includegraphics[scale=0.41]{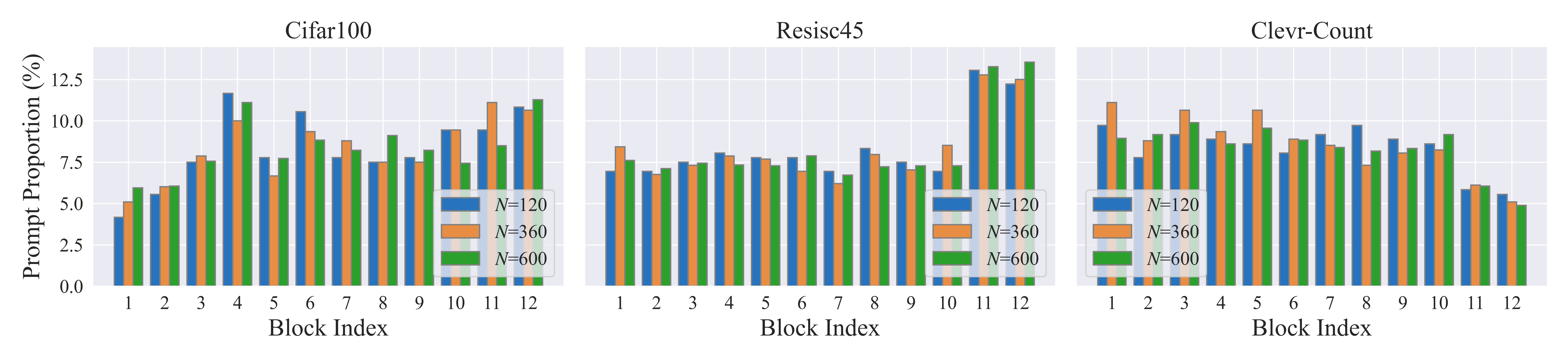}
  \vspace{-1.5em}
  \caption{\textbf{Visualization of prompt distributions learned by PRO-VPT} on VTAB-1k \textit{Natural} Cifar100, \textit{Specialized} Resisc45, and \textit{Structured} Clevr-Count with varying numbers of prompts. PRO-VPT effectively learns task-specific distributions.}
  \label{fig:observation1}
  \vspace{-1.2em}
\end{figure*}

\vspace{-0.4em}
\subsection{Observations on PRO-VPT} \label{sec:exp:understanding}
\vspace{-0.3em}
We design several experiments to thoroughly understand PRO-VPT. The key insights are concluded as follows:

\noindent \textbf{Observation 1.} 
\textit{PRO-VPT effectively learns task-specific prompt distributions.}
We visualize the learned distributions of PRO-VPT under different total numbers of prompts, as shown in Fig.~\ref{fig:observation1}. Notably, the distributions remain consistent across different prompt numbers but exhibit significant variations across different tasks. This indicates that the prompting importance of each block indeed varies by task, and PRO-VPT effectively captures and adapts to these task-specific distributions to better accommodate each task.

\begin{wrapfigure}{r}{0.194\textwidth}
  \centering
  \vspace{-1.3em}
  \hspace{-0.5em}
  \includegraphics[scale=0.38]{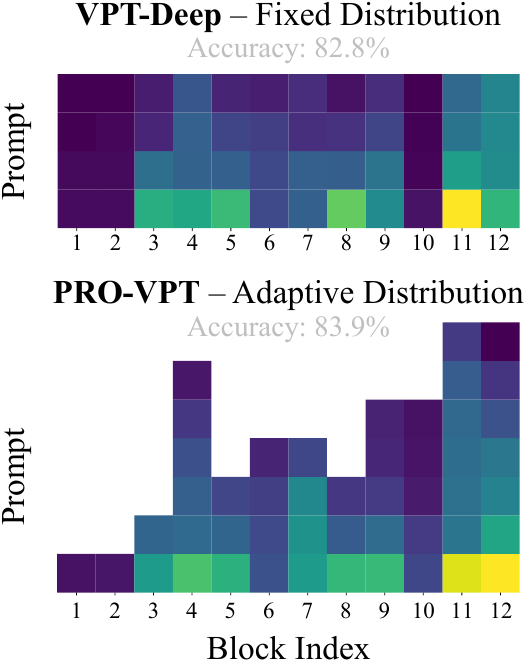}
  \vspace{-0.8em}
  \caption{\textbf{Visualization of prompt attention weights.}}
  \vspace{-2.0em}
  \label{fig:observation2}
\end{wrapfigure}

\noindent \textbf{Observation 2.} 
\textit{PRO-VPT effectively meets the block-specific prompt requirements via prompt relocation.}
We visualize the attention weights from representation features (\textit{i.e.}, [CLS] embeddings) to prompts across different blocks on Cifar100. Each attention weight represents the importance of a given prompt for its corresponding block. As depicted in Fig.~\ref{fig:observation2}, the attention weights in VPT reveal phenomena of prompt saturation (\textit{e.g.}, blocks 2 and 5) and prompt deficiency (\textit{e.g.}, blocks 4 and 12), confirming that different blocks exhibit significantly varying prompt demands.
However, due to its fixed prompt distribution, VPT lacks the flexibility to adapt to these diverse needs. Even pruning-based methods can only prune negative prompts from prompt-
saturated blocks while failing to address prompt-deficient blocks. In contrast, PRO-VPT effectively meets these block-specific prompt demands through adaptive prompt relocation. Such a fine-grained tuning strategy maximizes the effectiveness of the prompt distribution, thereby enhancing downstream task performance.

\begin{figure*}[!t]
  \centering
  \hspace{-0.75em} \includegraphics[scale=0.3475]{figs/provpt_stability.png}
  \vspace{-2.6em}
  \caption{\textbf{Cosine similarity of feature changes and the corresponding test accuracy.} The PR strategy has a minimal impact on features, and PRO-VPT demonstrates better stability than VPT in both representation and performance.}
  \label{fig:observation3_1}
  \vspace{-1.2em}
\end{figure*}

\vspace{0.3em}
\noindent \textbf{Observation 3.} 
\textit{PRO-VPT demonstrates enhanced stability and robustness.} 
One might be skeptical that runtime modifications of the prompt distribution could compromise stability. To investigate this, we evaluate both the representation stability and performance stability in PRO-VPT and VPT. Specific to feature stability, we measure cosine similarity between representations (\textit{i.e.}, [CLS] embeddings) from the previous epoch and after each stage of pruning, allocation, and tuning. This similarity reflects the extent of feature changes resulting from each stage. Fig.~\ref{fig:observation3_1} illustrates the similarity at different blocks and the corresponding test accuracy on Cifar100. Surprisingly, the orange and green lines indicate that the PR strategy has little impact on the learned features, as the features remain nearly unchanged after pruning and allocation. Moreover, PRO-VPT exhibits superior stability compared to VPT, both in terms of representation stability and performance stability. Additionally, as shown in Fig.~\ref{fig:observation3_2}, PRO-VPT demonstrates remarkable robustness to variations in prompt quantity compared to VPT, whereas the performance of VPT fluctuates significantly while PRO-VPT remains consistent. Overall, we argue that these benefits stem from our proposed \textit{optimization-based} approach, which effectively identifies the optimal road for relocation, thereby minimizing its impact on stability and enhancing robustness.

\vspace{-0.75em}
\section{Conclusion}
\label{sec:conclusion}
\vspace{-0.45em}
This paper investigates the ADO problem within VPT. Based on our experimental insights into the nested relationship between ADO and VPT, we formalize an ADO-VPT co-design framework through the lens of nested optimization. Upon this formalization, we propose a novel PRO-VPT algorithm, which iteratively optimizes the distribution by integrating the PR (prompt relocation) strategy with VPT. Across 26 datasets, we show our method's significant performance improvement over VPT with superior robustness.

\section*{Acknowledgements}
\vspace{-0.2em}
This work was supported in part by the National Natural Science Foundation of China under Grants 62376233, 62306181, 62476063, and 62306313; in part by the NSFC / Research Grants Council (RGC) Joint Research Scheme under Grant N\_HKBU214/21; in part by the General Research Fund of RGC under the Grant 12202622; in part by the RGC Senior Research Fellow Scheme under Grant SRFS2324-2S02; in part by the Guangdong Basic and Applied Basic Research Foundation under Grant 2024A1515010163; in part by the National Key Laboratory of Radar Signal Processing under Grant JKW202403; and in part by Xiaomi Young Talents Program.

{
    \small
    \bibliographystyle{ieeenat_fullname}
    \bibliography{main}
}

\clearpage
\appendix
\clearpage
\setcounter{page}{1}
\pagenumbering{roman}

\maketitlesupplementary

\setcounter{section}{0}
\setcounter{figure}{0}
\setcounter{table}{0}
\setcounter{equation}{0}

\renewcommand{\thetable}{\roman{table}}
\renewcommand{\thefigure}{\roman{figure}}
\renewcommand{\theequation}{\roman{equation}}
\renewcommand{\thealgocf}{\roman{algocf}}

\begin{figure*}[!ht]
  \centering
  \includegraphics[scale=0.245]{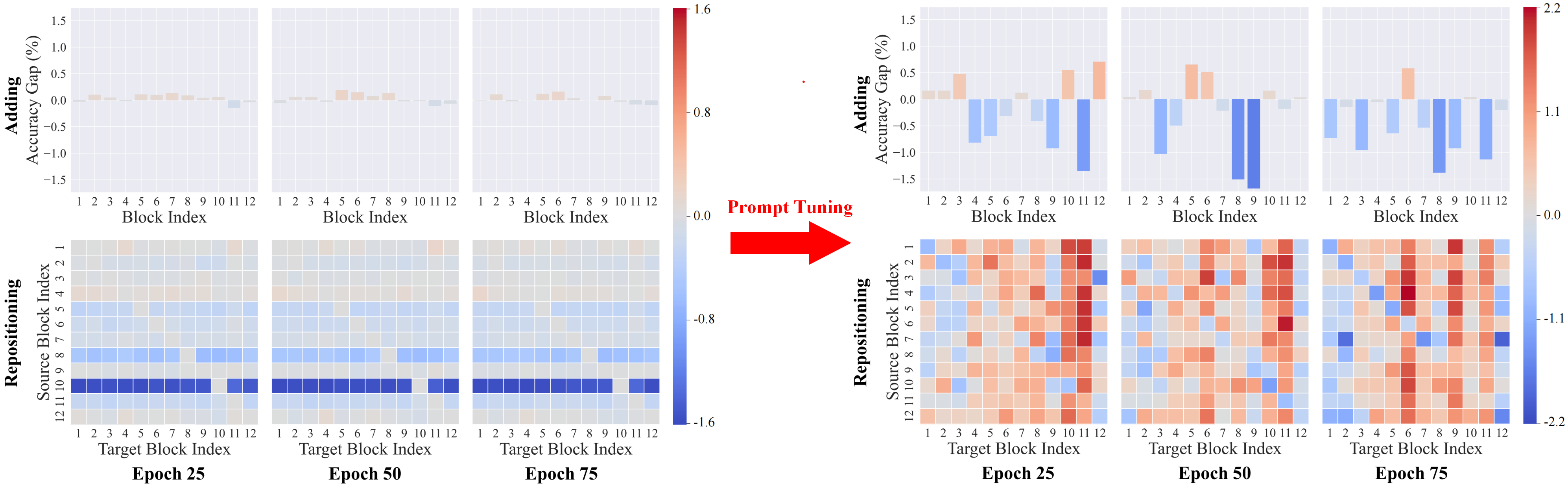}
  \vspace{-0.2em}
  \caption{\textbf{Detailed performance gaps from distribution adjustments} using prompts at epochs 25, 50, and 75, before and after prompt tuning. \textit{Left:} Performance gaps from adjustments before prompt tuning. \textit{Right:} Performance gaps from adjustments after prompt tuning. The effectiveness of distribution adjustments becomes apparent only after prompt tuning has been applied, and it also shifts with prompt updates.}
  \label{fig:appendix11}
\end{figure*}

This appendix presents further details and results that could not be included in the main paper due to space constraints. The content is organized as follows:
\begin{itemize}
    \item \S~\ref{sec:app_tmp} provides detailed explanations of the related technical concepts.
    \item \S~\ref{sec:app_tmp2} provides the detailed pseudo-code of PRO-VPT.
    \item \S~\ref{sec:app1} presents the complete results of distribution adjustments and offers an in-depth analysis of the underlying nature of ADO, which motivates the nested optimization formulation.
    \item \S~\ref{sec:app2} presents attempts at adding-based adjustments, which demonstrate significant instability.
    \item \S~\ref{sec:app3} explains why our proposed PR strategy relocates only a single prompt for each iteration.
    \item \S~\ref{sec:app4} details the derivation of the Taylor expansion for the idleness score.
    \item \S~\ref{sec:app5} provides more details of implementation.
    \item \S~\ref{sec:app6} presents complete results for VTAB-1k and FGVC as well as additional experimental results.
    \item \S~\ref{sec:app7} presents additional visualizations and analyses.
    \item \S~\ref{sec:app8} discusses the limitations of PRO-VPT and points out the potential direction for our future work.
\end{itemize}

\section{Detailed Explanations of Technical Concepts}  \label{sec:app_tmp}

To help readers better understand the relevant technical concepts mentioned in this paper, we provide a detailed explanation of the related approaches as follows:

\noindent \textbf{Vision Transformer.}
Given an input image $\mathbf{x}$, ViT~\cite{dosovitskiy2020vit, khan2022transformers, yang2024decoupling} first divides it into $n_e$ fixed-sized patches. Each patch is then embedded into $d$-dimensional latent space and combined with position encoding. The resulting set of patch embeddings is denoted as $\mathbf{E}_0\!=\! \{ \mathbf{e}_{0j} \!\in\! \mathbb{R}^d \}^{n_e}_{j=1}$. A learnable classification token $\mathbf{x}_0$ is then concatenated with these embeddings, forming the input sequence $[\mathbf{x}_{0}, \mathbf{E}_{0}]$. This sequence is then fed into a series of $L$ Transformer blocks $\{ B_i(\cdot) \}^L_{i=1}$ as follows:
\begin{equation}
    [\mathbf{x}_i, \mathbf{E}_i] = B_i\big([\mathbf{x}_{i-1}, \mathbf{E}_{i-1}]\big),~i = 1, 2, \ldots, L.
\end{equation}

\noindent \textbf{Visual Prompt Tuning.}
Given a set of trainable prompt tokens $\mathbf{P}$ and the prompted backbone $f_{\mathbf{P}}(\cdot)$, the overall objective of VPT~\cite{jia2022visual} is to optimize these prompts for effectively adapting the PVM to downstream tasks:
\begin{equation}
    \min_{\mathbf{P}}~\mathbb{E}_{(\mathbf{x}, y) \in \mathcal{T}_{tr}}\big[\mathcal{L}(f_\mathbf{P}(\mathbf{x}), y)\big].
\end{equation}
This formulation specifically corresponds to Eq.~\eqref{eq:prompt_tuning}.

Depending on how the prompt set $\mathbf{P}$ is distributed across the Transformer blocks, the standard VPT~\cite{jia2022visual} can be categorized into two variants, VPT-Shallow and VPT-Deep:

\textit{VPT-Shallow.}
The entire set of $p$ prompts, $\mathbf{P}\!=\! \{ \mathbf{p}_k \!\in\! \mathbb{R}^d \}^p_{k=1}$, is introduced merely in the first block. The shallow-prompted model is formulated as:
\begin{align}
    \hspace{-10.pt} [\mathbf{x}_1, \mathbf{Z}_1, \mathbf{E}_1] \!&=\! B_1\big([\mathbf{x}_0, \mathbf{P}, \mathbf{E}_0]\big), \\
    \hspace{-10.pt} [\mathbf{x}_i, \mathbf{Z}_i, \mathbf{E}_i] \!&=\! B_i\big([\mathbf{x}_{i-1}, \mathbf{Z}_{i-1}, \mathbf{E}_{i-1}]\big),~i \!=\! 2, 3, \ldots, L.
\end{align}

\textit{VPT-Deep.}
The prompt set $\mathbf{P}$ is uniformly distributed across all blocks, where each block $i$ is allocated a subset of $m$ prompts, $\mathbf{P}_i\!=\! \{ \mathbf{p}_{ik}\!\in\!\mathbb{R}^d \}^m_{k=1}$, with $\mathbf{P}\!=\! \bigcup^L_{i=1}\mathbf{P}_i$. The formulation of the deep-prompted model is as follows:
\begin{equation}
    \hspace{-4pt}
    [\mathbf{x}_i, \_, \mathbf{E}_i] = B_i\big([\mathbf{x}_{i-1}, \mathbf{P}_{i-1}, \mathbf{E}_{i-1}]\big),~i = 1, 2, \ldots, L. 
    \hspace{-3pt}
\end{equation}
where `$\_$' indicates that VPT-Deep does not preserve the output corresponding to the prompt tokens $\mathbf{P}_{i-1}$.

\noindent \textbf{Proximal Policy Optimization.} PPO~\cite{schulman2017proximal} is a widely used RL algorithm that can be applied to both discrete and continuous action spaces. Specifically, PPO-Clip updates policies by solving the following optimization problem:
\begin{equation}
    \theta_{k+1} = \mathop{\arg\min}\limits_{\theta}~\mathbb{E}_{s, a \sim \pi_{\theta_k}} \big[ L(s, a, \theta_k, \theta) \big],
\end{equation}
where $\pi$ is the policy, $\theta$ is the policy parameters, and $k$ is the $k$-th step. It typically takes multiple steps of SGD to optimize the clipped surrogate function $L(\cdot, \cdot, \cdot, \cdot)$:
{\footnotesize
\begin{align}
    &L(s, a, \theta_k, \theta) = \nonumber \\ 
    &\; \min \big( \frac{\pi_\theta(a\!\mid\! s)}{\pi_{\theta_k}(a\!\mid\! s)} A^{\pi_{\theta_k}} (s, a), \text{clip}(\frac{\pi_\theta(a\!\mid\! s)}{\pi_{\theta_k}(a\!\mid\! s)}, 1\!-\!\epsilon, 1\!+\!\epsilon)A^{\pi_{\theta_k}} (s, a) \big),
\end{align}
}
where $A^{\pi_{\theta_k}} (s, a)$ is the advantage estimator, and $\epsilon$ is the clip hyper-parameter. The clipping mechanism constrains the new policy to stay close to the old one, preventing excessively large policy updates that could degrade performance~\cite{lin2021end, schulman2017proximal}.

\begin{algorithm}[t]
    \SetCustomAlgoRuledWidth{0.45\textwidth}
    \SetAlgoLined
    \KwIn{pre-trained model $f$, number of epochs $T$, number of prompts $N$, learning rate $\eta$.}
    Initialize $N$ prompts $\mathbf{P}$ and distribute them to the model according to a uniform distribution $\mathcal{D}$. \\
    \For{$t=0,\ldots, T-1$} {
        \boxit{mypink}{8.52}
        Compute idleness scores $\{ \widehat{\mathcal{I}}_k \}^N_{k=1}$ by Eq.~\eqref{eq:idleness_score}. \\
        \If {$\max \widehat{\mathcal{I}}_k \!>\!0$} {
            \textbf{Prune} the negative prompt $\mathbf{p}_{k^{\!\ast\!}}$ with the maximum idleness score $\widehat{\mathcal{I}}_{k^{\!\ast\!}}$ to form $\mathcal{D}^{-}$. \\
            Compute the current state $s$. \\
            Compute the action $a^\ast \gets \text{PPO}(s)$.\\
            \textbf{Allocate} the idle prompt $\mathbf{p}_{k^{\!\ast\!}}$ to the $a^{\ast}$-th block to form $\mathcal{D}^{+}$. \\
        }
        \boxit{myblue}{1.16}
        Update the prompts as $\mathbf{P}^\prime \gets \mathbf{P} - \eta \cdot \mathbf{g}$.\\
        \boxit{mypink}{3.95}
        \If {$\max \widehat{\mathcal{I}}_k \!>\!0$} {
            Compute the reward $\hat{r}$ by Eq.~\eqref{eq:reward}.\\
            Update the policy networks within PPO based on $\hat{r}$.
        }
    }
    \caption{PRO-VPT}
    \label{alg:algo}
\end{algorithm}

\section{Pseudo-Code of PRO-VPT} \label{sec:app_tmp2}
We provide the detailed pseudo-code for PRO-VPT in Algo.~\ref{alg:algo}.
Initially, it generates $N$ prompts $\mathbf{P}$ and distributes them to the PVM according to a uniform distribution $\mathcal{D}$ (Line 1). In each epoch of PRO-VPT, it first calculates the estimated idleness scores $\{ \widehat{\mathcal{I}}_k \}^N_{k=1}$ using Eq.~\eqref{eq:idleness_score} (Line 3). If $\max \widehat{\mathcal{I}}_k \!>\! 0$, the PR process is triggered. This process begins by pruning the prompt $\mathbf{p}_{k^{\!\ast\!}}$ with the highest score $\widehat{\mathcal{I}}_{k^{\!\ast\!}}$, thereby constructing an intermediate distribution $\mathcal{D}^-$ (Line 5). The current state of the distribution $s$ is then computed, followed by determining the action $a^\ast$ based on PPO (Lines 6 and 7). Subsequently, the pruned prompt $\mathbf{p}_{k^{\!\ast\!}}$ is allocated to the $a^\ast$-th block, resulting in a relocated distribution $\mathcal{D}^+$ (Line 8). After completing the PR process, the prompts are optimized to $\mathbf{P}^\prime$ (Line 10). Additionally, if the PR process is activated during the epoch, it is necessary to calculate the estimated reward $\hat{r}$ using Eq.~\eqref{eq:reward} and update the policy networks in PPO accordingly (Lines 12 and 13).

\section{Comprehensive Analysis of the Underlying Nature behind ADO} \label{sec:app1}

Fig.~\ref{fig:appendix11} presents the complete results of Fig.~\ref{fig:ado_exp}. Specifically, we investigated the performance gaps from distribution adjustments applied to prompts at different epochs, both before and after prompt tuning, on VTAB-1k \textit{Natural} DTD using ViT-B/16. For the adding-based adjustment, we trained with a total of 59 prompts distributed uniformly across 12 Transformer blocks, resulting in one block containing 4 prompts while the others contained 5, then added a new prompt to the block with 4 prompts. For the repositioning-based adjustment, we trained with 60 prompts allocated uniformly, and then repositioned a single prompt. For fair comparisons, we averaged the results over five runs with different configurations, including variations in which block had the missing prompt as well as which specific prompt was repositioned.

The following provides a summary of the analysis and key findings presented in the main paper:

\noindent \textbf{Finding 3.} \textit{The effectiveness of distribution adjustments becomes apparent only after prompt tuning, suggesting that these adjustments can only be properly evaluated after tuning, thereby forming a nested relationship between ADO and VPT.} 
Comparing the left and right parts of Fig.~\ref{fig:appendix11}, we find that adjusting the distribution without prompt tuning leads to negligible changes or even degrades performance. In contrast, implementing prompt tuning after adjusting the distribution leads to significant performance changes, with well-chosen adjustments leading to notable improvements (\textit{e.g.}, an increase of up to 2.2 pp achieved by repositioning just a single prompt). To this end, the proper workflow for ADO and VPT should be established as `distribution adjustment $\rightarrow$ prompt tuning $\rightarrow$ adjustment evaluation', with prompt tuning nested within the distribution optimization process.

\noindent \textbf{Finding 2.} \textit{Adjustments in prompt distribution are influenced by the updated prompts themselves, underscoring the necessity of an iterative process that continuously refines the distribution over prompt updates.}
As demonstrated in the right portion of Fig.~\ref{fig:appendix11}, for prompts from epoch 25, we observe significant performance gains by adding a new prompt to the 10-th and 12-th blocks, as well as by repositioning one prompt to the 10-th and 11-th blocks. However, for prompts at epoch 75, the key improvements shifted to adding to the 6-th block and repositioning to the 6-th and 9-th blocks. To this end, the ADO-VPT co-design workflow should be established as an iterative process, which would be `distribution adjustment $\rightarrow$ prompt tuning $\rightarrow$ adjustment evaluation $\rightarrow$ new adjustment'.

\begin{figure*}[ht]
    \centering
    \begin{minipage}{0.4\linewidth}
        \centering
        \vspace{-0.5em}
        \includegraphics[scale=0.33]{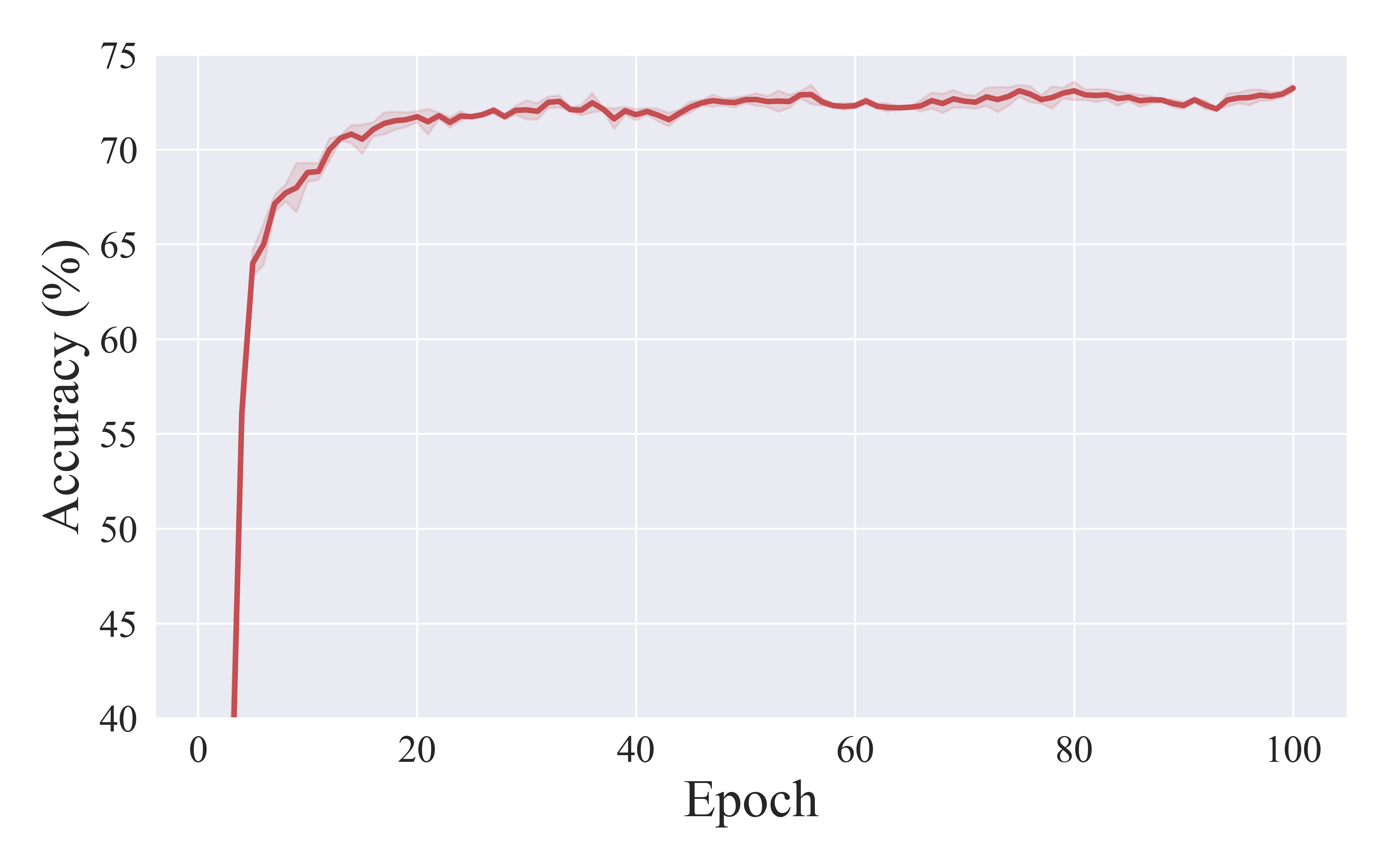}
        \vspace{-0.8em}
        \caption{\textbf{Convergence curve across different prompt-tuning epochs}, which corresponds to Fig.~\ref{fig:appendix11}.}
        \label{fig:appendix12}
    \end{minipage}
    \hspace{0.8em}
    \begin{minipage}{0.56\linewidth}
        \centering
        \includegraphics[scale=0.9]{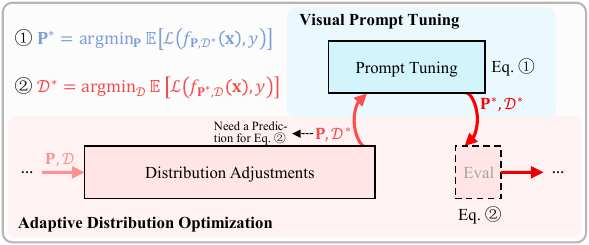}
        \caption{\textbf{Proper workflow of ADO-VPT co-design framework.} It is framed as an iterative process, with the VPT process nested within the ADO process.}
        \label{fig:appendix13}
    \end{minipage}
\end{figure*}

Additionally, to validate that the performance improvements from applying both distribution adjustments and prompt tuning mainly result from the distribution adjustments rather than the tuning itself, we present the respective prompt tuning accuracy curve in Fig.~\ref{fig:appendix12}. Clearly, the prompted model has already converged around epoch 20, with only slight performance fluctuations thereafter (less than 0.3 pp). This indicates that the performance differences observed in Fig.~\ref{fig:appendix11} (typically exceeding 0.3 pp) cannot be attributed solely to prompt tuning alone, but rather to the adjustments for the prompt distribution.

Building upon the underlying nature described in \textbf{Findings 2} and \textbf{3}, a proper workflow for ADO and VPT should be structured as an iterative nested process. Specifically, in each iteration, the process first adjusts the distribution to construct $\mathcal{D}^\ast$, then tunes the visual prompts to obtain $\mathbf{P}^\ast$. Based on $\mathcal{D}^\ast$ and $\mathbf{P}^\ast$, the effectiveness of the distribution adjustment is evaluated at the end of each iteration, marking the completion of the current cycle and the beginning of the next. Fig.~\ref{fig:appendix13} illustrates this co-design workflow for ADO and VPT. Formally, it can be expressed as a nested optimization problem as follows:
\begin{align}
    \label{eq:distribution_learning_app}
    \mathcal{D}^\ast &= \mathop{\arg\min}\limits_{\mathcal{D}}~\mathbb{E}_{(\mathbf{x}, y) \in \mathcal{T}_{tr}}\big[\mathcal{L}(f_{\mathbf{P}^\ast, \mathcal{D}}(\mathbf{x}), y)\big], \\
    \label{eq:prompt_tuning_app}
    \mathbf{P}^\ast &= \mathop{\arg\min}\limits_{\mathbf{P}}~\mathbb{E}_{(\mathbf{x}, y) \in \mathcal{T}_{tr}}\big[\mathcal{L}(f_{\mathbf{P}, \mathcal{D}^\ast}(\mathbf{x}), y)\big],
\end{align}
where the notations $\mathbf{P}^\ast$ and $\mathcal{D}^\ast$ correspond to those depicted in Fig.~\ref{fig:appendix13}.

Overall, the ADO problem is naturally formulated as a nested problem, grounded in its underlying nature.
This formulation is applicable to various distribution adjustment strategies (\textit{e.g.}, adding, repositioning, or even pruning) as well as different discrete optimization methods (\textit{e.g.}, evolutionary algorithms and reinforcement learning).
Notably, since distribution adjustments can only be evaluated at the end, a prediction for Eq.~\ref{eq:distribution_learning_app} is necessary for selecting an appropriate adjustment. To this end, it is natural to frame ADO as a RL problem, where the reward is predicted to determine the next action, and the effectiveness of that action is evaluated afterward.

\section{Inferior Performance of Adding-based Adjustments} \label{sec:app2}

\begin{figure}[b]
    \centering
    \includegraphics[scale=0.35]{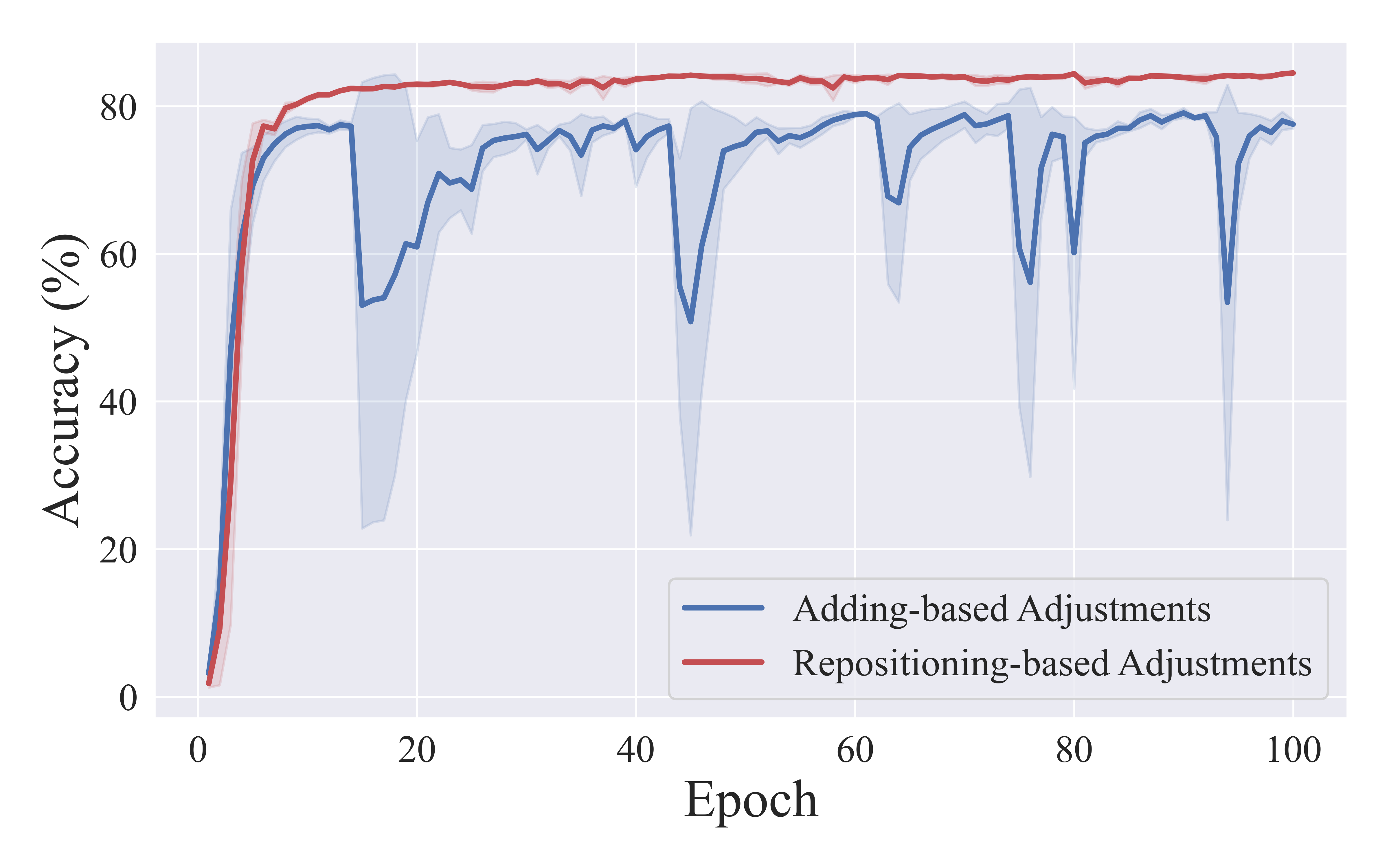}
    \vspace{-0.6em}
    \caption{\textbf{Convergence curves comparing adding-based and repositioning-based strategies.} The adding-based strategy exhibits significantly less stability.} \label{fig:appendix21}
\end{figure}

Following the nested optimization framework described in Eqs.~\eqref{eq:distribution_learning_app} and \eqref{eq:prompt_tuning_app}, we have also attempted to develop an adding-based strategy for ADO. Given that framing the ADO objective as a RL problem is a suitable choice (refer to \S~\ref{sec:method:pr} and \S~\ref{sec:app1}), we also address the adding-based ADO by leveraging RL, with the components of the Markov decision process specified as follows:

\textit{1) State.} We utilize the current prompt distribution as the state, denoted as $s = \mathcal{D}$. After adding a new prompt and performing prompt tuning, the state transitions to $s^\prime$.

\textit{2) Action.} Unlike prompt repositioning, which considers $L^2$ possible block arrangements, the adding-based ADO involves only $L$ potential adding operations. As a result, the adding-based ADO does not require decomposition for RL, and the action is straightforwardly represented as $a \in [L]$.

\textit{3) Reward.} Similar to the PR strategy, the reward is formulated based on Eq.~\eqref{eq:distribution_learning_app} as $r\!=\!\Delta \mathcal{L}(f_{\mathbf{P}, \mathcal{D}}, f_{\mathbf{P}^\prime, \mathcal{D}^\prime})$, where $\mathcal{D}^\prime$ denotes the updated prompt distribution after adding and $\mathbf{P}^\prime$ represents the tuned prompts. 

Similarly, we employ PPO for this RL problem to tackle the adding-based ADO objective, while adopting the overall framework illustrated in Fig.~\ref{fig:appendix13}.

Fig.~\ref{fig:appendix21} illustrates the performance comparison between the adding-based strategy and repositioning-based strategy (PRO-VPT) on the VTAB-1k \textit{Natural} Cifar100 dataset. It can be clearly observed that the adding-based strategy is significantly less stable and underperforms compared to the repositioning-based approach. We attribute this discrepancy to the undertraining of newly added prompts and potential conflicts with existing ones. Consequently, our work focuses on developing the repositioning-based strategy for ADO.

\section{Relocating One Prompt for Each Iteration} \label{sec:app3}

Here, we explain why we restrict the relocation process to operate on only a single prompt for each iteration. Since we frame the allocation step as a RL problem, we need to evaluate the effectiveness of each allocation decision for relocated prompts. Relocating more than one prompt simultaneously would necessitate multiple reward evaluations, substantially increasing task complexity and computational overhead. Therefore, we designed to relocate only one prompt per iteration to maintain simplicity and efficiency in the allocation process.

\section{Derivation of the Taylor Expansion for the Idleness Score} \label{sec:app4}

Based on the pruning objective in Eq.~\eqref{eq:pruning_goal}, the original idleness score is defined as:
\begin{equation}
    \mathcal{I}_k = \Delta \mathcal{L} (f_{\mathbf{P}, \mathcal{D}}, f_{\mathbf{P}, \mathcal{D}\mid d_k=0}),
\end{equation}
where $d_k=0$ indicates that the $k$-th prompt is pruned. Equivalently, this equation can be expressed as:
\begin{equation}
    \mathcal{I}_k = \Delta \mathcal{L} (f_{\mathbf{P}, \mathcal{D}}, f_{\mathbf{P}\mid \mathbf{p}_k=\mathbf{0}, \mathcal{D}}),
\end{equation}
where $\mathbf{p}_k=\mathbf{0}$ represents that the $k$-th prompt is a zero vector.

Let $p_{kj}\!\in\!\mathbf{p}_k$ denote as the prompt parameter. The difference in losses with and without the prompt parameter, \textit{i.e.}, the idleness score of $p_{kj}$, is given by:
\begin{equation}
    \mathcal{I}_{kj} = \Delta \mathcal{L} (f_{\mathbf{P}, \mathcal{D}}, f_{\mathbf{P}\mid p_{kj} = 0, \mathcal{D}}).
\end{equation}

\noindent Considering the entire prompt set as a concatenated vector $\mathbf{P}\!=\! \{ p_{00}, p_{01}, \ldots, p_{Nd} \}$, we are able to approximate $\mathcal{I}_{kj}$ in the vicinity of $\mathbf{P}$ by its first-order Taylor expansion:
\begin{equation}
\begin{split}
    \widehat{\mathcal{I}}_{kj} & \approx \mathbf{g}^T \Big( \mathbf{P} - \mathbf{P}\!\mid\!p_{kj} \!=\! 0 \Big) \\
    & \approx g_{kj}p_{kj},
\end{split}
\end{equation}
where $g_{kj}\!=\!\frac{\nabla \mathcal{L}}{\nabla p_{kj}}$ represents the element of the gradient $\mathbf{g}$.

For the idleness score of a prompt $\mathbf{p}_k\!=\!\{p_{kj}\}^d_{j=1}$, we can approximate it by summing the score of its individual parameters, as follows:
\begin{equation}
    \widehat{\mathcal{I}}_k \approx \sum_{j=1}^d g_{kj} p_{kj} \approx \mathbf{g}^T_k\mathbf{p}_k.
\end{equation}

To this end, we are able to efficiently approximate the idleness scores $\{\mathcal{I}_k\}^N_{k=1}$ by a single backpropagation pass, thereby avoiding the need to evaluate the idleness scores of all $N$ prompts individually.

\section{More Implementation Details} \label{sec:app5}

\textbf{Data Augmentation.}
Apart from data normalization, we resize the input images to 224$\times$224 pixels for VTAB-1k and apply a randomly resize crop to 224$\times$224 pixels and horizontal flipping for FGVC, as outlined in~\cite{jia2022visual, han2023e2vpt, steitz2024adapters, wang2025attention}. 

\begin{table}[!t]
    \centering
    \caption{\textbf{Hyper-parameters for VPT and PRO-VPT.}}
    \label{tab:app1}
    \scriptsize
    \begin{tabularx}{\linewidth}{@{}Xcc@{}}
    \toprule
    & VPT & PRO-VPT \\
    \midrule
    Batch size & 64 ($p\!\geq$100), 128 ($p\!<$100) & \hspace{4.75em}64\hspace{4.75em} \\
    Learning rate schedule & cosine decay & - \\
    Optimizer & \multicolumn{2}{c}{SGD} \\
    Optimizer momentum & \multicolumn{2}{c}{0.9} \\
    $base\_lr$ range & \multicolumn{2}{c}{\{50., 25., 10., 5., 2.5, 1., 0.5, 0.25, 0.1, 0.05\}} \\
    Weight decay range & \multicolumn{2}{c}{\{0.01, 0.001, 0.0001, 0.0\}} \\
    Drop rate & \multicolumn{2}{c}{0.1} \\
    Total epochs & \multicolumn{2}{c}{100} \\
    \bottomrule
    \end{tabularx}
    \vspace{0.9em}
\end{table}

\noindent \textbf{Training Hyper-Parameters.} 
Specific to training hyper-parameters, we largely adopt the same settings as depicted in VPT~\cite{jia2022visual}. Tab.~\ref{tab:app1} summarizes the hyper-parameter configurations comparing the experiments of VPT and our approach. Following~\cite{jia2022visual, han2023e2vpt}, we conduct a grid search on the validation set of each task to determine the optimal learning rate and weight decay; the learning rate is set as $base\_lr\!\times\! b/\text{256}$, where $b$ denotes the batch size and $base\_lr$ is selected from the range specified in Tab.~\ref{tab:app1}. Notably, PRO-VPT does not require specific-designed large learning rates as in~\cite{jia2022visual}. For all experiments conducted with our implementation, the results are averaged over three random seeds.

\noindent \textbf{PPO Hyper-Parameters.} 
We also detail the hyper-parameters of our PPO implementation for reproducibility. Both the actor and critic networks are two-layer MLPs with 64 hidden units per layer. The total number of parameters of the two policy networks is precisely 0.0136M. The learning rates are 0.0003 for the actor and 0.001 for the critic. Additionally, we set the discount factor to 1 and the clipping factor to 0.2.

\begin{table}[!t]
    \centering
    \caption{\textbf{Specifications of the VTAB-1k datasets.}}
    \label{tab:vtab_details}
    \footnotesize
    \sisetup{group-minimum-digits=4}
    \begin{tabularx}{\linewidth}{@{}l@{\enspace}XS[table-format=3.0]@{\quad}c@{\enspace\ }c@{\enspace\;}S[table-format=5.0]@{}}
    \toprule
    \multirow{2.4}*{Group} & \multirow{2.4}*{Task} & {\multirow{2.4}*{\#\,Classes}} & \multicolumn{3}{c}{Splits} \\\cmidrule{4-6}
     & & & {Train} & {Val} & {Test} \\
    \midrule
    \multirow{7}{*}{Natural} & CIFAR-100 & 100 & \multirow{7}*{\tablenum[table-format=3.0]{800}} & \multirow{7}*{\tablenum[table-format=3.0]{200}} & 10000 \\
    & Caltech-101 & 102 & & & 6084 \\
    & DTD & 47 & & & 1880 \\
    & Oxford Flowers & 102 & & & 6149 \\
    & Pets & 37 & & & 3669 \\
    & SVHN & 10 & & & 26032 \\
    & Sun397 & 397 & & & 21750 \\
    \midrule
    \multirow{4}{*}{Specialized} & Patch Camelyon & 2 & \multirow{4}*{\tablenum[table-format=3.0]{800}} & \multirow{4}*{\tablenum[table-format=3.0]{200}} & 32768 \\
    & EuroSAT & 10 & & & 5400 \\
    & RESISC45 & 45 & & & 6300 \\
    & Diabetic Retinopathy & 5 & & & 42670 \\
    \midrule
    \multirow{8}{*}{Structured} & CLEVR-Count & 8 & \multirow{8}*{\tablenum[table-format=3.0]{800}} & \multirow{8}*{\tablenum[table-format=3.0]{200}} & 15000 \\
    & CLEVR-Distance & 6 & & & 15000 \\
    & DMLab & 6 & & & 22735 \\
    & KITTI-Distance & 4 & & & 711 \\
    & dSprites-Location & 16 & & & 73728 \\
    & dSprites-Orientation & 16 & & & 73728 \\
    & smallNORB-Azimuth & 18 & & & 12150 \\
    & smallNORB-Elevation & 9 & & & 12150 \\
    \bottomrule
    \end{tabularx}
    \vspace{1.em}
\end{table}

\noindent \textbf{Datasets and Pre-Trained Backbones Specifications.} 
Tabs.~\ref{tab:vtab_details} and \ref{tab:fgvc_details} present the statistics of each task in VTAB-1k and FGVC \textit{w.r.t.} the number of classes and the number of images in the train, validation, and test splits. The tables are largely ``borrowed'' from~\cite{steitz2024adapters}. Moreover, Tab.~\ref{tab:model} provides the details of the pre-trained backbones used in this paper, which is largely ``borrowed'' from~\cite{jia2022visual}. 

\noindent \textbf{Reproducibility.}
PRO-VPT is implemented in Pytorch and timm. Experiments are conducted on NVIDIA A30-24GB GPUs. To guarantee reproducibility, our full implementation will be publicly released.

\section{More Experimental Results} \label{sec:app6}

\textbf{Complete Results for VTAB-1k.}
Tab.~\ref{tab:detailed_vtab1k_results} presents comprehensive results for VTAB-1k, using both ImageNet and Inception normalizations. Although several of the best pre-task results from other PEFT methods differ and exceed those listed in Tab.~\ref{tab:vtab1k_results}, PRO-VPT remains highly competitive, achieving a state-of-the-art average accuracy of 78.0\% among all evaluated methods.

\begin{figure}[!t]
    \centering
    \includegraphics[scale=0.39]{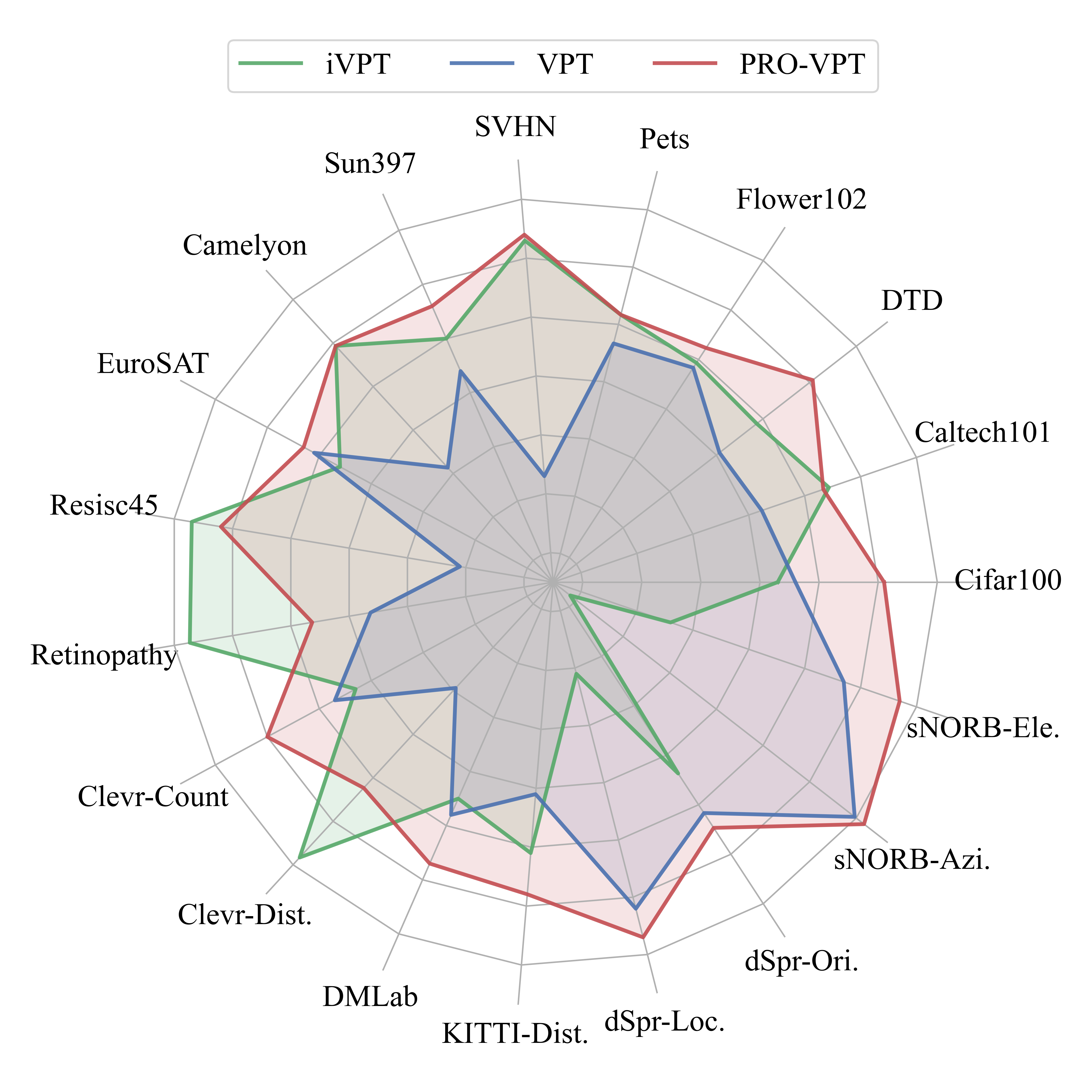}
    \vspace{-0.5em}
    \caption{\textbf{Comparison of prior state-of-the-art (iVPT), baseline (VPT), and our method (PRO-VPT).} Our approach subsumes two representative methods.} \label{fig:appendix61}
    \vspace{0.8em}
\end{figure}

Fig.~\ref{fig:appendix61} illustrates a comparison of prompt-based methods, including the previous state-of-the-art (iVPT), the baseline (VPT), and our proposal (PRO-VPT). All scores were normalized by $x_{\text{norm}}\!=\!x\!-\!x_{\text{mean}}$. It is evident that PRO-VPT outperforms both current leading methods, establishing a new state-of-the-art performance for prompt-based techniques.

\begin{table}[!t]
    \centering
    \caption{\textbf{Specifications of the FGVC datasets.} For datasets marked with *, we follow \cite{steitz2024adapters} to randomly sample train and validation splits since validation sets are not available from the original datasets.}
    \label{tab:fgvc_details}
    \footnotesize
    \sisetup{group-minimum-digits=4}
    \begin{tabularx}{\linewidth}{@{}lXS[table-format=3.0]@{\qquad}S[table-format=5.0]@{\quad\ }S[table-format=4.0]@{\quad\ }S[table-format=4.0]@{}}
    \toprule
    \multirow{2.4}*{Dataset} & & {\multirow{2.4}*{\#\,Classes}} & \multicolumn{3}{c}{Splits} \\\cmidrule{4-6}
    & & & {Train} & {Val} & {Test} \\
    \midrule
    CUB-200-2011*~\cite{wah2011caltech} & & 200 & 5394 & 600 & 5794 \\
    NABirds*~\cite{van2015building} & & 555 & 21536 & 2393 & 6084 \\
    Oxford Flowers~\cite{nilsback2008automated} & & 102 & 1020 & 1020 & 6149 \\
    Stanford Dogs*~\cite{dataset2011novel} & & 120 & 10800 & 1200 & 8580 \\
    Stanford Cars*~\cite{gebru2017fine} & & 196 & 7329 & 815 & 8041 \\
    \bottomrule
    \end{tabularx}
    \vspace{0.3em}
\end{table}

\begin{table}[!t]
    \centering 
    \caption{\textbf{Specifications of the pre-trained backbones.}}
    \label{tab:model} 
    \scriptsize
    \begin{tabularx}{\linewidth}{@{}Xlllll@{}}
    \toprule
    \hspace{0.5em}Backbone & \makecell[l]{Pre-trained\\Strategy} & \makecell[l]{Pre-trained\\Dataset} & \makecell[l]{Param\\(M)} & \makecell[l]{Feature\\dim $d$} & \makecell[l]{Pre-trained\\Model}\\
    \midrule
    \hspace{0.5em}ViT-B/16 & \multirow{3}{*}{Supervised} & \multirow{3}{*}{ImageNet-21k} & 85 & 768 & \href{https://storage.googleapis.com/vit_models/imagenet21k/ViT-B_16.npz}{checkpoint}\\
    \hspace{0.5em}ViT-L/16 &  &  & 307 & 1024 & \href{https://storage.googleapis.com/vit_models/imagenet21k/ViT-L_16.npz}{checkpoint}\\
    \hspace{0.5em}ViT-H/14 &  &  & 630 & 1280 & \href{https://storage.googleapis.com/vit_models/imagenet21k/ViT-H_14.npz}{checkpoint}\\
    \midrule
    \hspace{0.51em}Swin-B & Supervised & ImageNet-21k & 88 & 1024 & \href{https://github.com/SwinTransformer/storage/releases/download/v1.0.0/swin_base_patch4_window7_224_22k.pth}{checkpoint}\\
    \midrule
    \hspace{0.5em}ViT-B/16 & MAE & \multirow{2}{*}{ImageNet-1k}  & \multirow{2}{*}{85} & \multirow{2}{*}{768} & \href{https://dl.fbaipublicfiles.com/mae/pretrain/mae_pretrain_vit_base.pth}{checkpoint}\\
    \hspace{0.5em}ViT-B/16 & MoCo-v3 & &  &  & \href{https://dl.fbaipublicfiles.com/moco-v3/vit-b-300ep/linear-vit-b-300ep.pth.tar}{checkpoint\hspace{0.6em}}\\
    \bottomrule
    \end{tabularx}
\end{table}

\begin{table*}[!ht]
    \caption{\textbf{Comprehensive results on the VTAB-1k datasets.} Performance results are reported using both ImageNet normalization ($\circ$) or Inception normalization ($\bullet$), presented in \% after a complete training schedule with ViT-B/16 supervised pre-trained on ImageNet-21k. The best results of prompt-based methods and other PEFT approaches are highlighted in \textbf{bold}. \Lightning: Early-stopping based on the test set. $\dagger$: Lack of complete code or hyperparameter configurations for the method, hence results are reported as presented in the original paper. $^1$Average across the average accuracies of the VTAB-1k groups, following previous work.}
    \label{tab:detailed_vtab1k_results}
    \scriptsize
    \renewcommand*{\arraystretch}{0.95}
    \begin{tabularx}{\linewidth}{@{}+X-c@{\enspace\quad}-c@{\enspace}-c@{\enspace}-c@{\enspace}-c@{\enspace}-c@{\enspace}-c@{\enspace}-c@{\quad}-c@{\quad\enspace}-c@{\enspace}-c@{\enspace}-c@{\enspace}-c@{\quad}-c@{\quad\enspace}-c@{\enspace}-c@{\enspace}-c@{\enspace}-c@{\enspace}-c@{\enspace}-c@{\enspace}-c@{\enspace}-c@{\quad}-c@{\quad\enspace}-c@{}}
    \toprule
        & & \multicolumn{8}{c@{\enspace\quad}}{\textbf{Natural}} & \multicolumn{5}{c@{\enspace\quad}}{\textbf{Specialized}} & \multicolumn{9}{c@{\enspace\quad}}{\textbf{Structured}} \\
        \cmidrule(lr{2em}){3-10} \cmidrule(lr{2em}){11-15} \cmidrule(lr{2em}){16-24} \\ \addlinespace[-1em]
        & \rotatebox{90}{Param (M)} & \rotatebox{90}{Cifar100} & \rotatebox{90}{Caltech101} & \rotatebox{90}{DTD} & \rotatebox{90}{Flower102} & \rotatebox{90}{Pets} & \rotatebox{90}{SVHN}  & \rotatebox{90}{Sun397} & \rotatebox{90}{Group Avg.} & \rotatebox{90}{Camelyon} & \rotatebox{90}{EuroSAT}   & \rotatebox{90}{Resisc45}  & \rotatebox{90}{Retinopathy} & \rotatebox{90}{Group Avg.} & \rotatebox{90}{Clevr-Count} & \rotatebox{90}{Clevr-Dist.}  & \rotatebox{90}{DMLab} & \rotatebox{90}{KITTI-Dist.}  & \rotatebox{90}{dSpr-Loc.} & \rotatebox{90}{dSpr-Ori.}   & \rotatebox{90}{sNORB-Azi.}  & \rotatebox{90}{sNORB-Ele.} & \rotatebox{90}{Group Avg.} & \rotatebox{90}{Global Avg.$^1$}   \\
    \midrule
        Full $\circ$ & 85.8 & 68.9 & 87.7 & 64.3 & 97.2 & 86.9 & 87.4 & 38.8 & 75.9 & 79.7 & 95.7 & 84.2 & 73.9 & 83.4 & 56.3 & 58.6 & 41.7 & 65.5 & 57.5 & 46.7 & 25.7 & 29.1 & 47.6 & 69.0 \\
        Full $\bullet$ & 85.8 & 73.2 & 92.6 & 70.4 & 97.9 & 86.2 & 90.6 & 39.6 & 78.6 & 87.1 & 96.6 & {87.5} & 74.0 & 86.3 & 66.6 & 61.0 & 49.8 & 79.7 & 82.6 & 51.9 & 33.5 & 37.0 & 57.8 & 74.2 \\
        Linear $\circ$ & 0.04 & 63.4 & 85.0 & 63.2 & 97.0 & 86.3 & 36.6 & 51.0 & 68.9 & 78.5 & 87.5 & 68.6 & 74.0 & 77.2 & 34.3 & 30.6 & 33.2 & 55.4 & 12.5 & 20.0 & 9.6 & 19.2 & 26.9 & 57.7 \\
        Linear $\bullet$ & 0.04 & 78.1 & 88.1 & 69.0 & 99.1 & 90.0 & 36.0 & 56.9 & 73.9 & 79.8 & 90.7 & 73.7 & 73.7 & 79.5 & 32.4 & 30.5 & 35.9 & 61.9 & 11.2 & 26.2 & 14.3 & 24.5 & 29.6 & 61.0 \\
    \midrule
        LoRA $\bullet$ \cite{hu2022lora} & 0.29 & 83.0 & 91.7 & 71.6 & 99.2 & 90.9 & 83.8 & 56.7 & 82.4 & 86.2 & 95.7 & 83.5 & 71.9 & 84.3 & 77.7 & 62.3 & 49.0 & 80.2 & 82.2 & 51.7 & 31.0 & 47.0 & 60.1 & 75.6 \\
        FacT-TK$_{8}$ $\circ$ \cite{jie2023fact} & \textbf{0.05} & 70.3 & 88.7 & 69.8 & 99.0 & 90.4 & 84.2 & 53.5 & 79.4 & 82.8 & 95.6 & 82.8 & 75.7 & 84.2 & 81.1 & 68.0 & 48.0 & 80.5 & 74.6 & 44.0 & 29.2 & 41.1 & 58.3 & 74.0 \\
        $\text{FacT-TK}_{8}$ $\bullet$ \cite{jie2023fact} & \textbf{0.05} & 74.9 & 92.7 & 73.7 & 99.1 & 91.3 & 85.5 & \textbf{57.7} & 82.1 & 86.8 & 94.9 & 84.1 & 70.9 & 84.2 & 81.9 & 64.1 & 49.2 & 77.2 & 83.8 & 53.1 & 28.2 & 44.7 & 60.3 & 75.5 \\
        FacT-TK$_{\leq 32}$ $\circ$ \cite{jie2023fact} & 0.10 & 70.6 & 90.6 & 70.8 & 99.1 & 90.7 & 88.6 & 54.1 & 80.6 & 84.8 & 96.2 & 84.5 & 75.7 & 85.3 & 82.6 & \textbf{68.2}& 49.8 & 80.7 & 80.8 & 47.4 & 33.2 & 43.0 & 60.7 & 75.6 \\
        FacT-TK$_{\leq 32}$ $\bullet$ \cite{jie2023fact} & 0.10 & 74.6 & 93.7 & 73.6 & 99.3 & 90.6 & 88.7 & 57.5 & 82.6 & \textbf{87.6} & 95.4 & 85.5 & 70.4 & 84.7 & \textbf{84.3} & 62.6 & 51.9 & 79.2 & 85.5 & 52.0 & 36.4 & 46.6 & 62.3 & 76.5 \\
        Consolidator $\dagger$ \cite{hao2023consolidator} & 0.30 & 74.2 & 90.9 & 73.9 & \textbf{99.4} & 91.6 & \textbf{91.5} & 55.5 & 82.4 & 86.9 & 95.7 & 86.6 & \textbf{75.9} & 86.3 & 81.2 & \textbf{68.2} & 51.6 & \textbf{83.5} & 79.8 & 52.3 & 31.9 & 38.5 & 60.9 & 76.5 \\
        SSF \Lightning$\circ$ \cite{lian2022scaling} & 0.24 & 69.0 & 92.6 & \textbf{75.1} & \textbf{99.4}& 91.8 & 90.2 & 52.9 & 81.6 & {87.4} & 95.9 & 87.4 & 75.5 & \textbf{86.6} & 75.9 & 62.3 & 53.3 & 80.6 & 77.3 & 54.9 & 29.5 & 37.9 & 59.0 & 75.7 \\
        SSF \Lightning$\bullet$ \cite{lian2022scaling} & 0.24 & 61.9 & 92.3 & 73.4 & \textbf{99.4} & \textbf{92.0}& 90.8 & 52.0 & 80.3 & 86.5 & 95.8 & \textbf{87.5} & 72.8 & 85.7 & 77.4 & 57.6 & 53.4 & 77.0 & 78.2 & 54.3 & 30.3 & 36.1 & 58.0 & 74.6 \\
        SPT-Adapter \Lightning$\circ$ \cite{he2023sensitivity} & 0.23 & 72.9 & 93.2 & 72.5 & 99.3 & 91.4 & 84.6 & 55.2 & 81.3 & 85.3 & 96.0 & 84.3 & 75.5 & 85.3 & 82.2 & 68.0 & 49.3 & 80.0 & 82.4 & 51.9 & 31.7 & 41.2 & 60.8 & 75.8 \\
        SPT-Adapter \Lightning$\bullet$ \cite{he2023sensitivity} & 0.22 & 74.7 & 94.1 & 73.0 & 99.1 & 91.2 & 84.5 & 57.5 & 82.0 & 85.7 & 94.9 & 85.7 & 70.2 & 84.1 & 81.3 & 63.2 & 49.1 & 80.7 & 83.5 & 52.0 & 26.4 & 41.5 & 59.7 & 75.3 \\
        SPT-Adapter \Lightning$\circ$ \cite{he2023sensitivity} & 0.43 & 72.9 & 93.2 & 72.5 & 99.3 & 91.4 & 88.8 & 55.8 & 82.0 & 86.2 & 96.1 & 85.5 & 75.5 & 85.8 & 83.0 & 68.0 & 51.9 & 81.2 & 82.4 & 51.9 & 31.7 & 41.2 & 61.4 & 76.4 \\
        SPT-Adapter \Lightning$\bullet$ \cite{he2023sensitivity} & 0.43 & 74.9 & 93.2 & 71.6 & 99.2 & 91.1 & 87.9 & 57.2 & 82.2 & 87.0 & 95.4 & 86.5 & 72.4 & 85.3 & 81.1 & 63.2 & 50.3 & 80.2 & 84.4 & 51.4 & 31.5 & 42.2 & 60.5 & 76.0 \\
        Adapter+$_{r=16}$ $\bullet$ \cite{steitz2024adapters} & 0.35 & \textbf{83.7} & \textbf{94.2} & 71.5 & 99.3 & 90.6 & 88.2 & 55.8 & \textbf{83.3} & 87.5 & \textbf{97.0} & 87.4 & 72.9 & 86.2 & 82.9 & 60.9 & \textbf{53.7} & 80.8 & \textbf{88.4} & \textbf{55.2} & \textbf{37.3} & \textbf{46.9} & \textbf{63.3} & \textbf{77.6} \\

    \midrule
    \textcolor{\tblgray}{Prompt-based Methods:} & & & & & & & & & & & & & & & & & & & & & & & & \\
        VPT-Deep $\circ$ \cite{jia2022visual} & 0.60 & 78.8 & 90.8 & 65.8 & 98.0 & 88.3 & 78.1 & 49.6 & 78.5 & 81.8 & 96.1 & 83.4 & 68.4 & 82.4 & 68.5 & 60.0 & 46.5 & 72.8 & 73.6 & 47.9 & 32.9 & 37.8 & 55.0 & 72.0 \\
        VPT-Deep $\bullet$ \cite{jia2022visual} & 0.60 & 83.0 & 93.0 & 71.2 & 99.0 & 91.3 & 84.1 & 56.0 & 82.5 & 84.9 & 96.6 & 82.5 & 74.5 & 84.6 & 77.5 & 58.7 & 49.7 & 79.6 & 86.2 & 56.1 & 37.9 & 50.7 & 62.1 & 76.4 \\
        NOAH \Lightning$\circbullet$ \cite{zhang2024neural} & 0.43 & 69.6 & 92.7 & 70.2 & 99.1 & 90.4 & 86.1 & 53.7 & 80.2 & 84.4 & 95.4 & 83.9 & 75.8 & 84.9 & \textbf{82.8} & \textbf{68.9} & 49.9 & \textbf{81.7} & 81.8 & 48.3 & 32.8 & 44.2 & 61.3 & 75.5 \\
        SPT-Deep $\dagger$ \cite{wang2024revisiting}  & \textbf{0.22} & 79.3 & 92.6 & \textbf{73.2} & \textbf{99.5} & 91.0 & 89.1 & 51.2 & 82.3 & 85.4 & \textbf{96.8} & 84.9 & 74.8 & 85.5 & 70.3 & 64.8 & \textbf{54.2} & 75.2 & 79.3 & 49.5 & 36.5 & 41.5 & 58.9 & 75.6 \\
        iVPT $\dagger$ \cite{zhou2024ivpt} & 0.60 & 82.7 & \textbf{94.2} & 72.0 & 99.1 & \textbf{91.8} & 88.1 & 56.6 & 83.5 & \textbf{87.7} & 96.1 & \textbf{87.1} & \textbf{77.6} & \textbf{87.1} & 77.1 & 62.6 & 49.4 & 80.6 & 82.1 & 55.3 & 31.8 & 47.6 & 60.8 & 77.1 \\
        PRO-VPT (\textbf{ours}) & 0.61 & \textbf{84.5} & 94.1 & \textbf{73.2} & 99.4 & \textbf{91.8} & \textbf{88.2} & \textbf{57.2} & \textbf{84.1} & \textbf{87.7} & \textbf{96.8} & 86.6 & 75.5 & 86.7 & 78.8 & 61.0 & 50.6 & 81.3 & \textbf{86.7} & \textbf{56.4} & \textbf{38.1} & \textbf{51.7} & \textbf{63.1} & \textbf{78.0} \\
    \bottomrule
    \end{tabularx}
\end{table*}

\begin{table}[t]
    \caption{\textbf{Comprehensive results on the FGVC datasets.} Performance results are reported as the highest of ImageNet normalization ($\circ$) or Inception normalization ($\bullet$), presented in \% after a complete training schedule with ViT-B/16 supervised pre-trained on ImageNet-21k. The best results of prompt-based methods and other PEFT approaches are highlighted in \textbf{bold}. $\dagger$: Lack of complete code or hyperparameter configurations for the method, hence results are reported as presented in the original paper.}
    \label{tab:detailed_fgvc_results}
    \footnotesize
    \begin{tabularx}{\columnwidth}{@{}+X-c@{\enspace\quad}-c@{\enspace}-c@{\enspace}-c@{\enspace}-c@{\enspace}-c@{\enspace\quad}-c@{}}
    \toprule
        & \rotatebox{90}{Param (M)} & \rotatebox{90}{CUB200} & \rotatebox{90}{NABirds} & \rotatebox{90}{Oxford Flowers} & \rotatebox{90}{Stanford Dogs }& \rotatebox{90}{Stanford Cars} & \rotatebox{90}{Global Avg.} \\
    \midrule
    Full $\circ$ & 86.0 & 87.3 & 82.7 & 98.8 & 89.4 & 84.5 & 88.5 \\
    Full $\bullet$ & 86.0 & 88.0 & 81.5 & 99.2 & 85.6 & 90.6 & 89.0 \\
    Linear $\circ$ & 0.18  & 85.3 & 75.9 & 97.9 & 86.2 & 51.3 & 79.3 \\
    Linear $\bullet$ & 0.18  & 88.9 & 81.8 & 99.5 & 92.6 & 52.8 & 83.1 \\
    \midrule
    SSF $\circ$ \cite{lian2022scaling} & 0.39 & 89.5 & \textbf{85.7} & 99.6 & 89.6 & \textbf{89.2} & 90.7 \\
    SSF $\bullet$ \cite{lian2022scaling} & 0.39 & 88.9 & 85.0 & 99.6 & 88.9 & 88.9 & 90.3 \\
    SPT-Adapter $\dagger$ \cite{he2023sensitivity} & 0.40 & 89.1 & 83.3 & 99.2 & 91.1 & 86.2 & 89.8 \\
    SPT-LoRA $\dagger$ \cite{he2023sensitivity} & 0.52 & 88.6 & 83.4 & 99.5 & 91.4 & 87.3 & 90.1 \\
    Adapter+ $\bullet$ \cite{steitz2024adapters} & \textbf{0.34} & \textbf{90.4} & 85.0 & \textbf{99.7} & \textbf{92.6} & 89.1 & \textbf{91.4} \\
    \midrule
    \textcolor{\tblgray}{Prompt-based Methods:} & & & & & & & \\
    VPT-Deep $\circ$ \cite{jia2022visual} & 0.85 & 88.5 & 84.2 & 99.0 & 90.2 & 83.6 & 89.1 \\
    VPT-Deep $\bullet$ \cite{jia2022visual} & 0.85 & 90.1 & 83.3 & 99.6 & 90.3 & 85.0 & 89.7 \\
    SPT-Deep $\dagger$ \cite{wang2024revisiting} & \textbf{0.36} & \textbf{90.6} & \textbf{87.6} & \textbf{99.8} & 89.8 & 89.2 & 91.4 \\
    iVPT $\dagger$ \cite{zhou2024ivpt} & 0.41 & 89.1 & 84.5 & 99.5 & 90.8 & 85.6 & 89.9 \\
    PRO-VPT (\textbf{ours}) & 0.86 & \textbf{90.6} & 86.7 & 99.7 & \textbf{91.8} & \textbf{89.6} & \textbf{91.7} \\
    \bottomrule
    \end{tabularx}
\end{table}

\begin{table}[t]
    \caption{\textbf{Generalizability study on detection and segmentation tasks.} Results are presented on two instances: COCO val2017 and ADE20k.}
    \label{tab:more_tasks}
    \scriptsize
    \renewcommand*{\arraystretch}{1.0}
    \setlength{\tabcolsep}{1.4mm}
    \begin{threeparttable}
    \begin{tabularx}{\linewidth}{@{}Xcccccccc@{}}
    \toprule
    & \multicolumn{6}{c}{COCO with Mask R-CNN} & \multicolumn{2}{c}{ADE20k with SETR} \\ \addlinespace[-0.1em]
    \cmidrule(lr){2-7} \cmidrule(lr){8-9} \\ \addlinespace[-1.1em]
    & AP$^b$ & AP$^b_\text{50}$ & AP$^b_\text{75}$ & AP$^m$ & AP$^m_\text{50}$ & AP$^m_\text{75}$ & mIoU-SS & mIoU-MS \\
    \midrule
    VPT-Deep & 33.8 & 57.6 & 35.3 & 32.5 & 54.5 & 33.9 & 39.1 & 40.1 \\
    PRO-VPT & \textbf{34.6} & \textbf{58.6} & \textbf{36.1} & \textbf{33.4} & \textbf{55.5} & \textbf{34.7} & \textbf{40.0} & \textbf{41.0} \\
    \bottomrule
    \end{tabularx}
    \begin{tablenotes}
    \item[1] AP$^b$ and AP$^m$ are the average precision for objective detection and instance segmentation.
    \item[2] mIoU-SS and mIoU-MS are single- and multi-scale inference of semantic segmentation.
    \end{tablenotes}
    \end{threeparttable}
    \vspace{0.8em}
\end{table}

\noindent \textbf{Complete Results for FGVC.}
Tab.~\ref{tab:detailed_fgvc_results} presents comprehensive results for FGVC, utilizing both ImageNet and Inception normalizations. The best pre-task results remain consistent with those in Tab.~\ref{tab:fgvc_results}, and PRO-VPT demonstrates superior performance on large-scale fine-grained datasets.

\begin{figure}[!t]
    \centering
    \vspace{-1em}
    \includegraphics[scale=0.39]{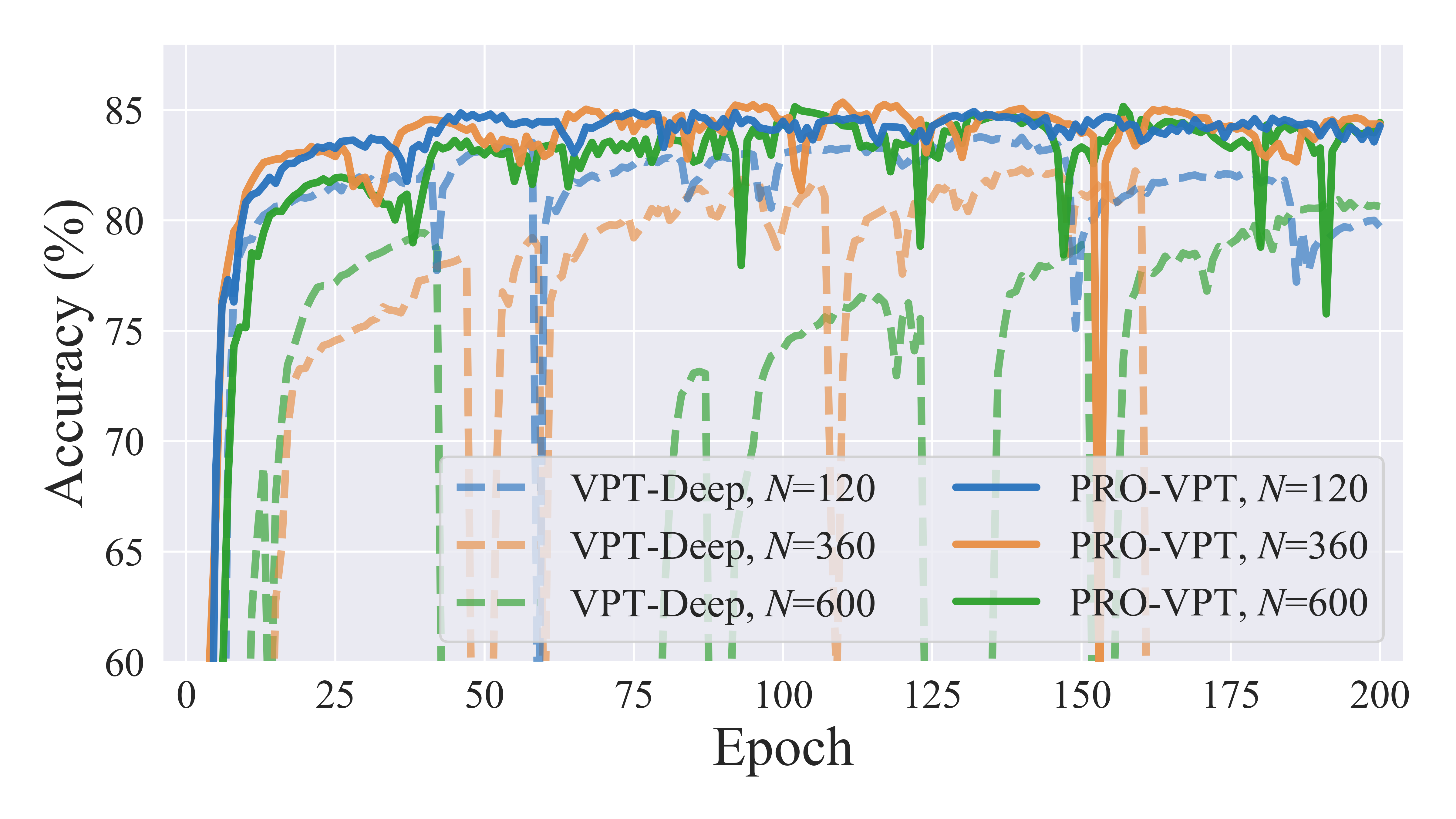}
    \vspace{-1.2em}
    \caption{\textbf{Convergence curves comparing VPT and PRO-VPT} with varying numbers of prompts. PRO-VPT exhibits superior robustness.} \label{fig:observation3_2}
    \vspace{0.8em}
\end{figure}

\noindent \textbf{Generalizability Study on Detection and Segmentation Tasks.}
We also conduct experiments on a broader range of downstream tasks, including object detection, instance segmentation, and semantic segmentation. Specifically, we evaluate object detection and instance segmentation performance on the COCO dataset~\cite{lin2014microsoft}, using Mask R-CNN~\cite{he2017mask} with a Swin-T backbone pre-trained on ImageNet-1k. For the semantic segmentation task, we use the ADE20k dataset~\cite{zhou2017scene} and adopt SETR-PUP~\cite{zheng2021rethinking} with a ViT-B/16 backbone pre-trained on ImageNet-21k. As shown in Tab.~\ref{tab:more_tasks}, the results demonstrate that PRO-VPT consistently outperforms VPT across both detection and segmentation tasks, highlighting its superior generalizability.

\noindent \textbf{Convergence Curves with Different Prompt Numbers.} 
Fig.~\ref{fig:observation3_2} illustrates the convergence curves under varying total numbers of prompts on VTAB-1k \textit{Natural} Cifar100, with training extended to 200 epochs. Notably, VPT exhibits high sensitivity to the number of prompts, as discussed in~\cite{wang2024revisiting, huang2024cvpt}. An improper prompt number can lead to significant instability during convergence and result in inferior performance. In contrast, PRO-VPT demonstrates remarkable robustness to variations in prompt quantity. Although it is slightly affected during the initial convergence phase, our method ultimately converges to consistent performance. This suggests that learning the optimal distribution enables a more nuanced calibration of tuning intensity at each block, thereby enhancing robustness.

\begin{figure}[!t]
  \centering
  \includegraphics[scale=0.33]{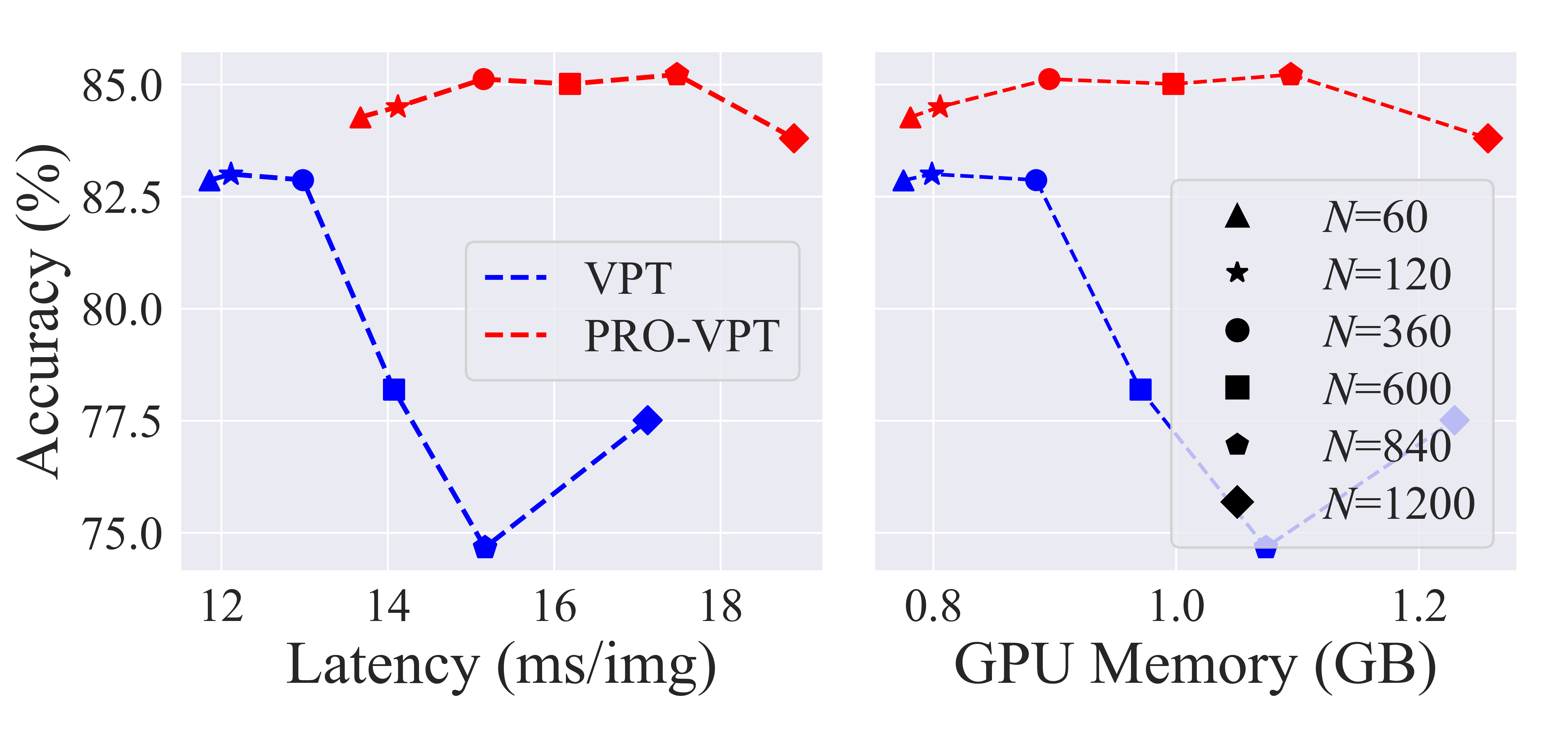}
  \vspace{-0.8em}
  \caption{\textbf{Cost comparisons} with different prompt numbers. We report the latency (ms/img) and the GPU memory usage (GB). The increased training time associated with PRO-VPT is justified by the considerable enhancements in both performance and robustness.
  }
  \label{fig:cost}
    \vspace{0.8em}
\end{figure}

\noindent \textbf{Cost Analysis.}
As detailed in \S~\ref{sec:method}, estimating the expected objectives effectively avoids additional computation, thereby significantly enhancing the efficiency of our method. In particular, the extra training cost of our method primarily stems from the computation of policy networks in PPO. However, since the policy networks are merely lightweight MLPs, this additional cost is relatively low. Specifically on VTAB-1k, the average latency for VPT and PRO-VPT is 13.37 ms/img and 15.55 ms/img (\textbf{1.16}$\mathbf{\times}$). Furthermore, we evaluate latency and GPU usage with varying prompt numbers on VTAB-1k \textit{Natural} Cifar100 in Fig.~\ref{fig:cost}. Although PRO-VPT does introduce some extra training time, the performance and robustness improvements justify this cost, and the increase in GPU memory usage is marginal.

\section{Visualization and Analysis} \label{sec:app7}

\begin{figure}[!t]
  \centering
  \hspace{-0.5em}
  \includegraphics[scale=0.58]{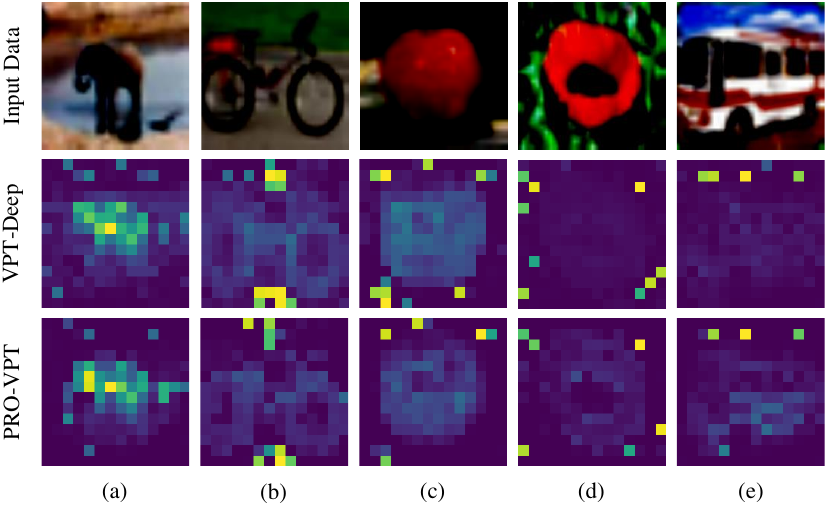}
  \caption{\textbf{Visualization of attention maps}. PRO-VPT exhibits more focused and precise attention with fewer artifacts.}
  \label{fig:attn2}
  \vspace{0.8em}
\end{figure}

\begin{figure*}[!ht]
  \centering
  \hspace{-0.5em}\includegraphics[scale=0.91]{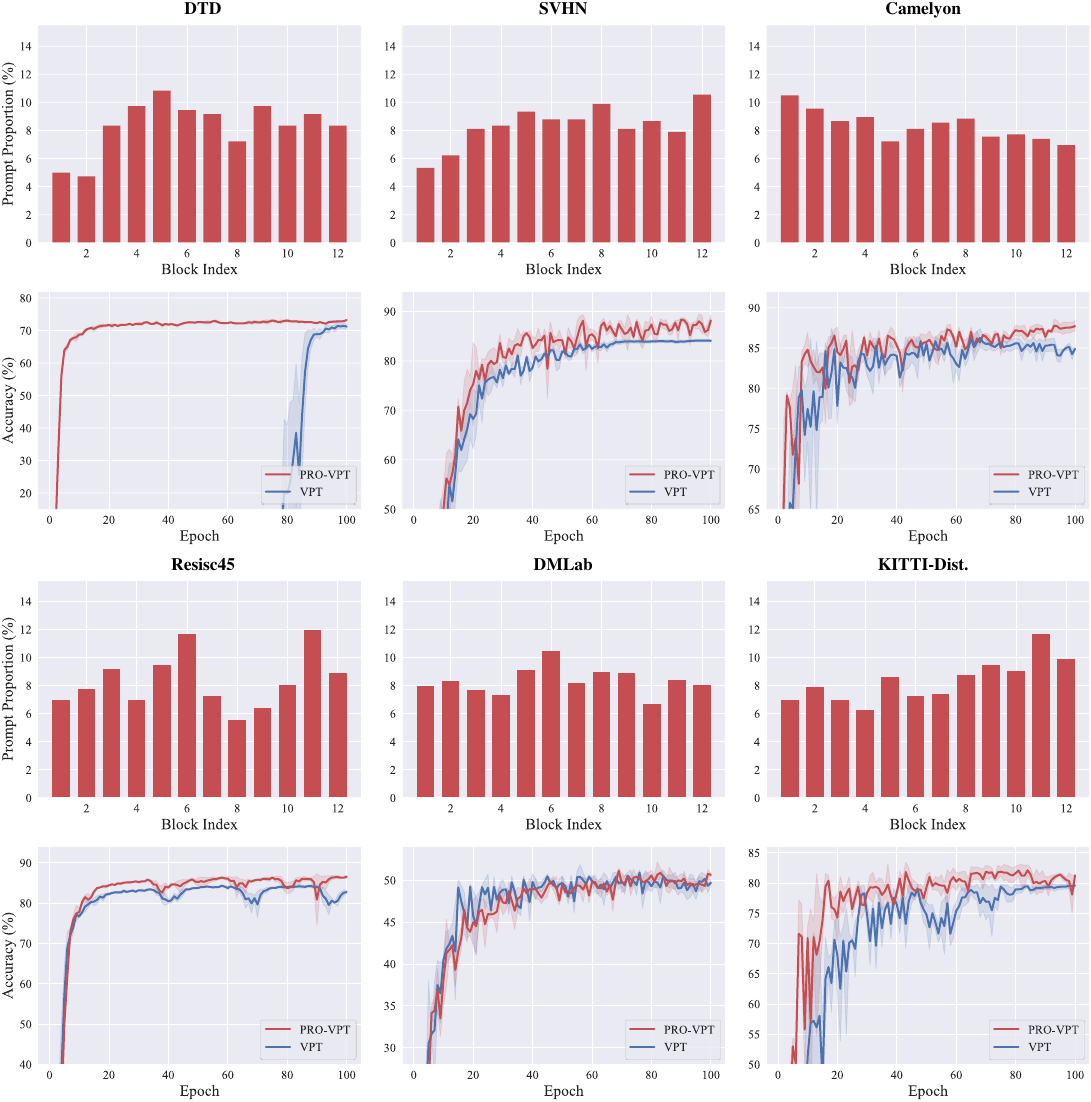}
  \caption{\textbf{Visualization of the prompt distributions learned by PRO-VPT and the accuracy curves compared to VPT} on the VTAB-1k datasets: \textit{Natural} DTD, SVHN; \textit{Specialized} Camelyon, Resisc45; and \textit{Structured} DMLab, KITTI-Dist.}
  \label{fig:appendix71}
\end{figure*}

\noindent \textbf{Attention Maps.} 
We visualize the attention maps between [CLS] and image patches on VTAB-1k \textit{Natural} Cifar100. As shown in Figs.~\ref{fig:attn2}(a)-(c), while VPT successfully focuses on the object, its attention exhibits significant artifacts~\cite{darcet2024vision} and, more critically, remains scattered. For example, in Fig.~\ref{fig:attn2}(a), VPT shows certain attention on the lake, which is actually part of the background. In contrast, PRO-VPT demonstrates more focused and accurate attention with fewer artifacts. Furthermore, as illustrated in Figs.~\ref{fig:attn2}(d) and (e), VPT appears to struggle with effectively concentrating on the object, whereas PRO-VPT maintains its ability to focus on the object.

\noindent \textbf{Learned Distributions and Accuracy Curves.} Fig.~\ref{fig:appendix71} illustrates the learned distributions in PRO-VPT as well as the accuracy curves in comparison to VPT, based on 100 training epochs. Empirical results from more datasets further reinforce that the prompting importance for each block is inherently task-dependent. Moreover, the comparison of accuracy curves between PRO-VPT and VPT, particularly on the SVHN, Camelyon, and Resisc45 datasets, reveals that PRO-VPT still exhibits an upward trend in accuracy during the late training stages. This highlights the effectiveness of learning the optimal distribution for visual prompts, which unlocks their full potential and maximizes downstream performance.

\section{Limitation and Future Work} \label{sec:app8}

Despite demonstrating promising performance and enhanced robustness, our approach still has certain limitations. 
First, as noted in~\cite{wang2024revisiting, huang2024cvpt}, prompt-based methods are significantly sensitive to the total number of prompts. Although our method improves robustness and mitigates this issue to some extent (refer to \S~\ref{sec:exp:understanding}), there is still room for further refinement. In particular, while PRO-VPT maintains robust performance within a certain range (\textit{e.g.}, 120-600 initialized prompts as in Fig.~\ref{fig:observation3_2}), its performance still deteriorates when the number of prompts deviates significantly from the optimal value. Consequently, our method still employs the optimal total number of prompts per task from~\cite{jia2022visual}, resulting in relatively high parameter counts as in VPT. 
Second, the constraint of relocating one prompt per iteration (see \S~\ref{sec:app3}) restricts our method to relocating only limited prompts within certain epochs. This particularly affects tasks requiring high prompt numbers, hindering PRO-VPT's ability to learn optimal distributions. For instance, the changes from initial uniform to learned prompt distributions are relatively less pronounced for Camelyon and DMLab as shown in Fig.~\ref{fig:appendix71}, which require 100 initialized prompts per block. Nevertheless, it is worth noting that PRO-VPT still delivers considerable performance enhancements on these datasets. Addressing these limitations would further enhance the efficiency and applicability of our approach in diverse downstream tasks.

Moreover, while this paper focuses on calibrating the distribution of prompts, the PRO-VPT framework is applicable to most block-wise PEFT approaches, \textit{e.g.}, adjusting the pre-block dimension of subspaces (\textit{i.e.}, the rank distribution) in adapter-based methods. An essential future direction deserving of further investigation is integrating our approach with other PEFT methods, developing a universal framework for fine-grained robust PEFT.

\end{document}